\documentclass[1p,times,authoryear]{elsarticle}

\usepackage{amsmath}

\usepackage{makecell}
\usepackage{float}
\usepackage{enumitem}
\usepackage{amsmath}
\usepackage{subfigure}
\usepackage{float}
\usepackage{cite}
\usepackage{svg}
\usepackage{amsmath}
\usepackage{graphics}
\usepackage{graphicx}
\usepackage{amssymb}
\usepackage{algorithm}
\usepackage{subcaption}  
\usepackage{booktabs}    
\usepackage{multirow}    
\usepackage{adjustbox}
\usepackage{mathrsfs}
\usepackage{siunitx}
\usepackage{booktabs}
\usepackage{bm}
\usepackage{bbm}
\usepackage{makecell}
\usepackage{multirow}
\usepackage{multicol}
\usepackage{float}
\usepackage{url}
\usepackage{amsfonts}
\usepackage{textcomp}
\usepackage{supertabular}
\usepackage{array}
\usepackage{hyperref}
\usepackage{amsfonts}

\graphicspath{{images/}}
\usepackage[pagewise]{lineno} 
\usepackage{xcolor}
\usepackage{soul} 
\usepackage{changes}
\usepackage{amssymb}

\begin{document}
\makeatletter
\def\ps@pprintTitle{%
  \let\@oddhead\@empty
  \let\@evenhead\@empty
  \let\@oddfoot\@empty
  \let\@evenfoot\@empty}
\makeatother
\begin{frontmatter}

\title{AgriFM: A Multi-source Temporal Remote Sensing Foundation Model for Agriculture Mapping}

\author[a]{Wenyuan Li}
\author[a]{Shunlin Liang\corref{corresponding author}}
\ead{shunlin@hku.hk}
\author[b]{Keyan Chen}
\author[a]{Yongzhe Chen}
\author[a]{Han Ma}
\author[a]{Jianglei Xu}
\author[a]{Yichuan Ma}
\author[a]{{Yuxiang Zhang}}
\author[a]{Shikang Guan}
\author[c]{Husheng Fang}
\author[b]{Zhenwei Shi}

\cortext[corresponding author]{Corresponding author: Shunlin Liang.}

\address[a]{Jockey Club STEM Lab of Quantitative Remote Sensing, Department of Geography, The
University of Hong Kong, Hong Kong, China}
\address[b]{Department of Aerospace Intelligent Science and Technology, School of Astronautics, Beihang University, Beijing, China}
\address[c]{School of Remote Sensing and Information Engineering, Wuhan University, China}

\begin{abstract}

Climate change and population growth intensify the demand for precise agriculture mapping to enhance food security. Such mapping tasks require robust modeling of multi-scale spatiotemporal patterns from fine field textures to landscape context, and from short-term phenology to full growing-season dynamics. Existing methods often process spatial and temporal features separately, limiting their ability to capture essential agricultural dynamics. While transformer-based remote sensing foundation models (RSFMs) offer unified spatiotemporal modeling ability, most of them remain suboptimal: they either use fixed windows that ignore multi-scale crop characteristics or neglect temporal information entirely. To address these gaps, we propose AgriFM, a multi-source, multi-temporal foundation model for agriculture mapping. AgriFM introduces a synchronized spatiotemporal downsampling strategy within a Video Swin Transformer backbone, enabling efficient handling of long and variable-length satellite time series while preserving multi-scale spatial and phenological information. It is pre-trained on a globally representative dataset comprising over 25 million samples from MODIS, Landsat-8/9, and Sentinel-2 with land cover fractions as pre-training supervision. AgriFM further integrates a versatile decoder specifically designed to dynamically fuse multi-source features from different stages of backbone and accommodate varying temporal lengths, thereby supporting consistent and scalable agriculture mapping across diverse satellite sources and task requirements. It supports diverse tasks including agricultural land mapping, field boundary delineation, agricultural land use / land cover mapping, and specific crop mapping (e.g., winter wheat and paddy rice) with difference data sources. Comprehensive evaluations show that AgriFM consistently outperforms existing deep learning models and general-purpose RSFMs across multiple agriculture mapping tasks. Codes and models are available at \url{https://github.com/flyakon/AgriFM} and \url{https://glass.hku.hk}.

\end{abstract}


\end{frontmatter}



\section{Introduction}
\label{sec:introduction}

The dual challenges of rapid population growth and intensifying climate change impacts have elevated global food security to a pressing priority \citep{wheeler2013climate, singh2023climate}. Satellite remote sensing has become indispensable for addressing this challenge, providing effective tools for large-scale agricultural monitoring and precise crop mapping \citep{van2023worldcereal, fang2024comprehensive, chen5090896first}. Recent technological advances have enhanced both the spatial and temporal resolution of Earth observation data \citep{yuan2020deep, ma2025first,liang2021global,jiang2023hi}, creating unprecedented opportunities for detailed {agriculture mapping}. However, it simultaneously demands more sophisticated analytical methods capable of extracting meaningful agricultural intelligence from complex spatiotemporal patterns.

The growing demands for accurate {agriculture mapping (e.g. agricultural land mapping, agricultural land use / land cover mapping, and crop mapping)} have spurred significant methodological evolution, beginning with phenology-based approaches.
They utilize the distinct physical attributes of various crops, including their differing reflectance across spectral bands and their unique phenological traits at various growth phases. Examining and detecting the alterations throughout the crop growth cycle can efficiently differentiate various crops \citep{qiu2022maps, dong2020early, qiu2017winter, liang2024mapping, ashourloo2022new}. Qiu et al. \citep{qiu2022maps} introduced an innovative approach for generating yearly 500-m MODIS-derived national maps of various cropping systems in China, employing phenology-based mapping algorithms and pixel purity thresholds, resulting in an overall accuracy of 89\%. Ashourloo et al. \citep{ashourloo2022new} introduced a novel phenology-based approach utilizing Sentinel-2 time-series data for the efficient differentiation of wheat and barley across extensive regions, attaining an overall accuracy exceeding 76\%.
The issue with these methods is that they require setting different thresholds based on the mapping task and study area, which significantly  affects their applicability. For complex scenarios and tasks, this may lead to erroneous or even failed mapping results.

Machine learning methods \citep{yin2020monitoring, yang2023automated, xuan2023mapping, van2023worldcereal} have, to a certain extent, resolved the issue of adaptability. They establish the relationship between satellite spectral data and {agriculture land use / land covers}, utilizing algorithms  such as Random Forest \citep{breiman2001random} and Support Vector Machine (SVM) \citep{hearst1998support}. The effectiveness of machine learning largely depends on obtaining sufficient ground truth data, which is challenging in many scenarios. 
Van Tricht et al. \citep{van2023worldcereal} introduced the WorldCereal, a global applicable, seasonally updated crop and irrigation mapping product, which leverages the Random Forest algorithm. {Yin et al.} \citep{yin2020monitoring} developed a novel method based on the Random Forest algorithm to map the extent and timing of abandoned cropland using the Landsat time series and tested this approach in 14 diverse global study regions.
Despite the wide application in {agriculture mapping} tasks, machine learning methods struggle to simultaneously understand the spatial and temporal information from satellite observations. As a result, they often underperform in complex tasks and lack sufficient processing capacity for large-scale remote sensing data.

Deep learning, as a subset of machine learning methods,  has proven to be a potent tool in processing and understanding big data, propelling significant advancements across diverse fields \citep{li2021geographical,miller2024deep,li5029097asiawheat}.  
Deep learning models include Convolutional Neural Networks (CNNs) \citep{he2016deep}, Long Short-Term Memory (LSTM) \citep{yu2019review}, and Transformers \citep{vaswani2017attention}. Waldner et al. \citep{waldner2020deep} proposed a data-driven method using ResUNet-a, a deep convolutional neural network with a fully connected UNet backbone, for accurate and scalable extraction of field boundaries from satellite data. 
As CNNs cannot process temporal remote sensing images, some studies have explored the use of 3D CNNs \citep{hara2018can} or LSTMs to provide temporal features for these tasks. Gallo et al. \citep{gallo2023season} proposed an innovative method using 3D Convolutional Neural Networks to process Sentinel-2 time series data, enabling in-season and dynamic crop mapping with real-time updates. Furthermore, some studies have incorporated LSTM to capture additional temporal information, all of which have yielded promising results \citep{turkoglu2021crop, russwurm2023end, barriere2024boosting}. 

Recently, transformer models have achieved breakthrough progress in computer vision \citep{ zhang2024vision}, remote sensing image processing \citep{chen2024rsprompter}, and earth observation tasks \citep{li2024deepphysinet}. Remote Sensing Foundation Models (RSFMs), built on the basis of transformers, are considered a superior paradigm for handling multi-source temporal remote sensing data \citep{zhou2023comprehensive}. Vision Transformer (ViT) \citep{dosovitskiy2020image} and Swin Transformer \citep{liu2021swin,liu2022video} are two prominent transformer architectures utilized in the construction of foundation models. RSFMs typically use a large volume of remote sensing data for pre-training, and then a small amount of labeled data for fine-tuning to accomplish specific tasks \citep{tan2023promises, zhu2024foundations, zhou2024towards, lu2024ai, zhang2024foundation}. The key to pre-training is to help the network learn representations from a large amount of remote sensing data \citep{cong2022satmae, reed2023scale}. Pre-training methods include masked imaging modeling (MIM) \citep{hondru2024masked} and contrastive learning (CL) \citep{chen2020simple, he2020momentum, hondru2024masked}. In some cases, if a large amount of labeled remote sensing data can be obtained, supervised pre-training is also an effective pre-training method \citep{feng2021rethinking,chen2025dynamicvis}.           

{Transformer-based RSFMs have shown strong potential in extracting spatiotemporal features from remote sensing data. This ability to model temporal sequences end-to-end and integrate multi-source data makes them particularly promising for agricultural applications, where capturing complete crop growth cycles is critical for accurate mapping}\citep{tseng2025galileo,sumbul2025smarties}. 
Several pioneering studies have begun exploring RSFMs for crop mapping tasks. Fang et al. \citep{fang5138538generating} developed a rice mapping method using NASA-IBM's Prithvi foundation model \citep{Prithvi-100M-preprint} with Harmonized Landsat and Sentinel-2 (HLS30) data \citep{claverie2018harmonized} and  successfully generated rice distribution maps across Monsoon Asia. {Similarly, Qin et al.} \citep{qin2025spatiotemporal} proposed a spatiotemporal masking strategy for pretraining a spatiotemporal collaborative learning network (STCLN) to extract informative representations for crop mapping. 

While these approaches demonstrate promising results, they predominantly rely on Vision Transformer (ViT) architectures. This preference stems from ViT's straightforward temporal data processing and relatively lower computational demands during masked image modeling pretraining. However, ViT's inherent downsampling mechanism and fixed spatiotemporal patch windows may adversely affect crop mapping performance. Previous research \citep{chen2017deeplab, huang2020unet} has proven pixel-wise classification tasks require high spatial feature fidelity, yet ViT's patch embedding operation typically employs aggressive and fixed downsampling (e.g., 16×16 patches), potentially compromising subtle spatial-temporal differences. Although Fang and Qi's implementations incorporate specialized modifications to mitigate these limitations, such adaptations increase methodological complexity while limiting generalizability.

In contrast, Swin Transformer's hierarchical feature extraction capability, resembling CNN's multi-scale processing, appears better suited for agriculture mapping tasks. Like U-Net architectures, Swin Transformer can effectively fuse features across scales to enhance spatial precision \citep{cao2022swin}. However, existing Swin-based RSFMs remain scarce and typically ignore temporal information, severely limiting their applicability for crop monitoring. While Video Swin Transformer \citep{liu2022video} inherently supports joint spatiotemporal processing, few existing RSFMs have yet adopted it as the backbone architecture. 

To address these gaps, this paper first establishes the necessity of simultaneous hierarchical spatiotemporal feature extraction for {agriculture  mapping}, leading to the development of a modified Video Swin Transformer \citep{liu2022video} architecture where temporal down-sampling is  synchronized with spatial down-sampling operations.  {This modified backbone enables efficient unified processing of long and variable-length satellite time series while preserving critical multi-scale spatial patterns and phenological dynamics.} We then develop AgriFM, a multi-source temporal remote sensing foundation model specifically designed for {agriculture mapping}.  

AgriFM leverages temporally rich data streams from MODIS, Landsat-8 / 9, and Sentinel-2 satellites, pre-trained on
a global representative dataset with supervision from global land cover products. 
The pretraining dataset comprises over 25 million globally sampled images from MODIS (250m \& 500m), Landsat-8/9 (30m), and Sentinel-2 (10m \& 20m). Leveraging the synchronized spatio-temporal downsampling strategy, we dynamically sample 3-32 frames from each satellite source during every pre-training iteration. By adaptively adjusting the downsampling ratios across temporal scales, feature maps of consistent dimensions can be obtained while accommodating variable sequence lengths. The foundation model thus learns to extract robust representations from diverse temporal contexts, effectively capturing multi-scale phenological patterns across different agricultural monitoring scenarios.

Unlike {other} RSFMs relying on MIM or CL for pretraining, we incorporate land cover fraction from GLC\_FCS30D data \citep{zhang2024glc_fcs30d} for supervised pretraining. Specifically, we extract image-level land cover fractions within each sample area as regression targets, optimizing with L1 loss. This approach follows established practices demonstrating land cover priors' effectiveness \citep{li2022geographical,wang2024multi}. To mitigate the impact of supervision noises caused by inaccuracies in reference products, we implement an auxiliary teacher network with exponential moving average updates, effectively guiding the model toward robust feature learning while filtering out unreliable supervision signals.

AgriFM incorporates a versatile decoder to dynamically fuse extracted multi-scale spatiotemporal representations, supporting diverse agriculture  mapping tasks, {including agricultural land mapping, field boundary delineation,  agricultural land use/ land cover mapping and specific crop  mapping (e.g., winter wheat and paddy rice).}
For systematic evaluation, we compare AgriFM with three representative approaches: (1) ViT-based RSFMs (Prithvi \citep{Prithvi-100M-preprint}, SatMAE \citep{cong2022satmae}, {Galileo \citep{tseng2025galileo}, SMARTIES \citep{sumbul2025smarties}}), (2) Swin-based models without temporal processing (PIS \citep{PIS}, {GFM \citep{GFM}}), and (3) conventional deep learning methods (CNNs / LSTMs). Experimental results demonstrate AgriFM's consistent superiority, 
particularly in two crucial aspects: preserving fine-grained spatial details essential for field-level mapping, and effectively modeling long-term temporal patterns critical for crop type analysis,  where existing methods show notable limitations. AgriFM achieves unified processing of heterogeneous data sources, variable temporal sequences, and diverse mapping tasks within a single network architecture. This comprehensive method overcomes the traditional need for task-specific model adaptations while delivering robust performance across various tasks.

\section{Methodology and Data}
\begin{figure*}[!htb]
\centering
\includegraphics[width=\linewidth]{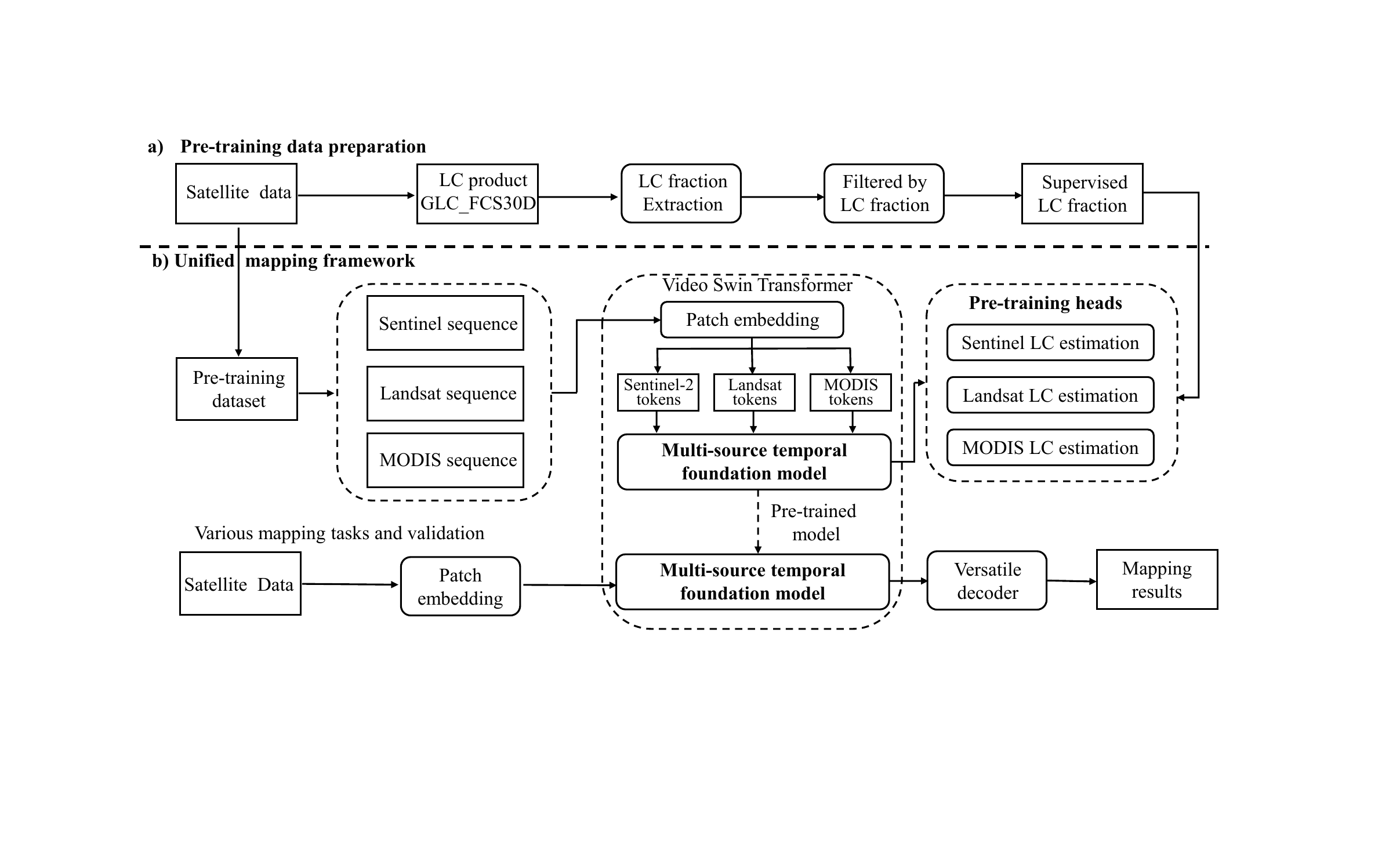}
\caption{{Flowchart outlining the AgriFM: a) The initial phase involves the extraction of pre-training supervision from geographical priors (land cover products) and the assembly of an extensive pre-training dataset, b) The subsequent phase entails the pre-training of the multi-source temporal foundation model and construction of unified mapping framework.}}
\label{fig:flowchart}
\end{figure*}

In this section, we will delve into the implementation specifics surrounding the AgriFM. The entire flowchart is graphically represented in Figure \ref{fig:flowchart} and can be systematically divided into two primary phases. The first phase involves the creation of a large-scale pre-training dataset and the extraction of land cover fractions to serve as pre-training supervision. The second phase pertains to the development of our multi-source temporal foundation model, along with details of the pre-training process. Based on the foundation mode AgriFM, a unified mapping framework is achieved through the construction of a versatile decoder and subsequent fine-tuning with labels.

\subsection{Pre-training Data Preparation}
We build a global pre-training dataset composed of multi-source temporal satellite data and land cover fractions from global land cover product GLC\_FCS30D \citep{zhang2024glc_fcs30d}. 

\subsubsection{Multi-source Satellite Data for Pre-training}
\begin{figure*}[!htb]
\centering
\includegraphics[width=\linewidth]{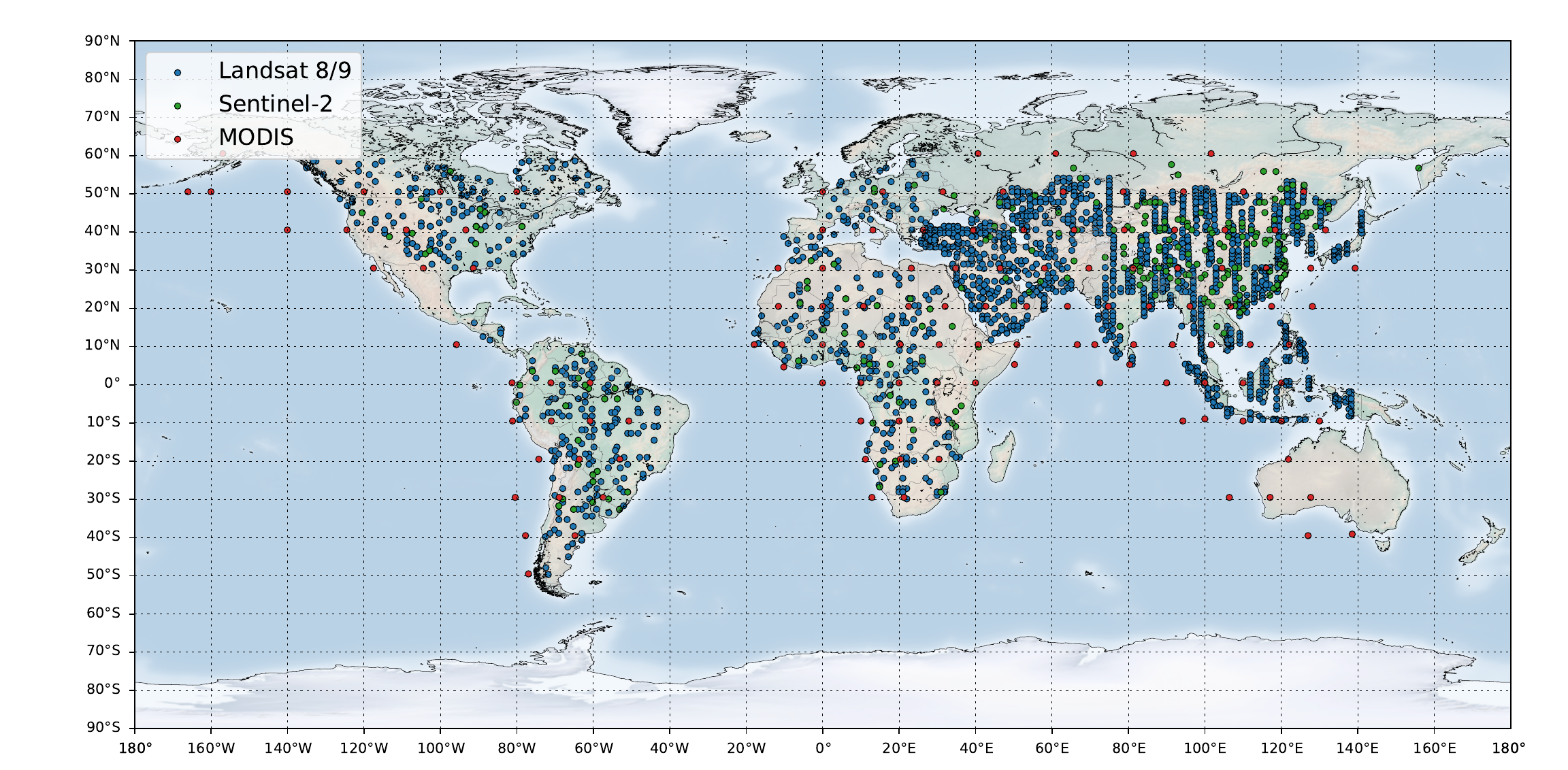}
\caption{{The spatial distribution of pre-training samples collected on a global scale from Sentinel-2, Landsat-8/9 and MODIS.}}
\label{fig:dist_pretraining}
\end{figure*}

Satellite data, serving as the input for the foundation models during pre-training, are indiscriminately collected on a global scale from Sentinel-2, Landsat-8/9 and MODIS. The spatial distribution of these pre-training samples is depicted in Figure \ref{fig:dist_pretraining}.

The MODIS data from the Terra satellite include the surface reflectance products MOD09A1 and MOD09Q1 with the temporal resolution of 8 days and the spatial resolutions of   500 m and 250 m, respectively. The MOD09Q1 product comprises the first two bands at 250m, and we extract bands 3-7 at 500m from the MOD09A1 product and then re-sample them to 250 meters. The MODIS data selected for pretraining are  from  2020 to 2022.

The Landsat surface reflectance data at 30m are extracted from the NASA Harmonized Landsat and Sentinel-2 (HLS) project \citep{claverie2018harmonized}. This is a NASA initiative aimed at creating a harmonized surface reflectance product from the Operational Land Imager (OLI) and Multi-Spectral Instrument (MSI) aboard the Landsat-8/9 (L30) and Sentinel-2A/B (S30) remote sensing satellites, respectively. The initiative involves adapting the original projection of Landsat 8/9 to correspond with the projection of Sentinel-2. For the pre-training phase, we globally collect Landsat 8/9 data from HLS L30, with a 30-meter resolution and six bands: Blue, Green, Red, NIR, SWIR 1, and SWIR 2 in 2022. The HLS L30 dataset enhances the native 16-day revisit cycle of individual Landsat satellites to a nominal 4-7 day effective temporal resolution through dual-satellite synergy (Landsat 8/9) and overlapping orbit observations, particularly crucial for capturing rapid crop phenological transitions.

We use the Sentinel-2 data's four bands at 10-m and six bands at 20 m. These bands encompass visible and near-infrared wavelengths, specifically advantageous for differentiating diverse types of vegetation, identifying water bodies, and monitoring urban areas. They  include Blue, Green, Red, Red-Edge 1, Red-Edge 2, Red-Edge 3, NIR B08, NIR B08A, SWIR 1, and SWIR 2, and all of them are re-sampled to 10m spatial resolution. All samples are collected in 2021 and 2022, and with a 5-day temporal resolution.

\subsubsection{Land Cover Product for Pre-training Supervision}

We utilize the GLC\_FCS30D \citep{zhang2024glc_fcs30d} global land cover product for the extraction of land cover fractions for each pre-training sample. These extracted fractions are then used as the pre-training supervision for the foundational model.

GLC\_FCS30D is a product of global land cover maps spanning from 1985 to 2022. These maps, which have been generated using Landsat data at a 30m resolution, contain three levels of land cover and a total of 35 finely classified types. The basic classification types we use for pre-training encompass cropland, forest, shrubland, grassland, wetland, water bodies, bare land, and impervious surfaces.

During the pre-training phase, we segment the original satellite data into 224 x 224 sized images. For each image, we determine the land cover distribution within the region it covers, and calculate the fractions accordingly \citep{li2021geographical}. These land cover fractions will then serve as pre-training supervision during the pre-training phase.
For any input remote sensing image $\mathbf{I}$, we first define its geographic footprint. We denote the coordinates of the upper left corner as $[lon_{min}, lat_{max}]$ and the lower right corner as $[lon_{max}, lat_{min}]$. The spatial extent $\Omega$ is then defined as the rectangular region spanning from $[lon_{min}, lat_{max}]$ to $[lon_{max}, lat_{min}]$. From the land cover product $\mathcal{L}$, we retrieve the corresponding classification map $\mathbf{L}$ within $\Omega$:
\begin{equation}
\mathbf{L} = \mathcal{L}(i,j), \quad \forall (i,j) \in \Omega,
\end{equation}
where $\mathbf{L}$ denotes the land cover map and  $\mathbf{L}_{i,j} \in {1,2,…,K}$ denotes the land cover class at pixel $(i,j)$, with $K$ being the total number of land covers. {From this step, we extract all pixels in region $\Omega$ with upper left coordinates $[lon_{min},lat_{max}]$ and lower right coordinates $[lon_{max},lat_{min}]$} In our implementation, we focus on the eight most frequent land covers in the dataset: (1) cropland, (2) forest, (3) shrubland, (4) grassland, (5) wetland, (6) water, (7) bare land, and (8) urban areas. Pixels belonging to other classes are aggregated as background.

The land cover fraction for the k-th land cover class within $\Omega$ is denoted as $p_k$ and is computed as:
\begin{equation}
p_k = \frac{1}{N} \sum_{i=1}^H \sum_{j=1}^W \mathbb{I}(\mathbf{L}_{i,j} = k).
\end{equation}
$N=H\times W$ is the total number of pixels in $\Omega$. $\mathbb{I}(\dots)$ is the indicator function (1 if true, 0 otherwise). $H$ and $W$ represent the spatial dimensions of the classification map.

The final land cover fraction feature vector is constructed as:
\begin{equation}
\mathbf{p} = [p_0, p_1, p_2, \dots, p_8]^T \in \mathbb{R}^9,
\end{equation}
where $p_1$ to $p_8$ represent the fractional coverage of their respective land cover classes, while $p_0$ denotes the collective fraction of all other categories not explicitly listed, serving as the background or no-data component. 

We employ land cover fractions as supervision instead of the commonly used Masked Image Modeling (MIM) approach for three compelling reasons. First, existing studies \citep{li2021geographical} have demonstrated that land cover guidance enables models to acquire valuable geographic priors during pre-training. While the precise mechanistic impact remains challenging to quantify, empirical evidence confirms this constraint significantly improves pre-training efficiency and stability. Methodologically, supervised pre-training aligns with the proven paradigm of early ImageNet pre-training models \citep{ImageNet}, whose effectiveness has been extensively validated. The predominance of MIM largely stems from the scarcity of natural image annotations - a limitation irrelevant to our case, as geo-referenced remote sensing data automatically provides spatially aligned land cover information through geographic coordinates at near-zero annotation cost. This unique advantage allows us to leverage authoritative land cover products while maintaining full scalability.

\subsubsection{Pre-training Dataset}
After the pre-processing, we have constructed a multi-source temporal remote sensing pre-training dataset, which is derived from MODIS, Landsat 8/9, and Sentinel-2 satellite data at  250m\&500m, 30m, and 10m\&20m spatial resolution. To meet the input requirements of deep learning models, we randomly select these satellite data and crop the original satellite data into images of 224 x 224 pixels. For each image, a sequence of time series data is also generated with a minimum length of 16. Land cover fractions are computed to serve as pre-training supervision and the image with only background land cover is abandoned. 

The pre-training dataset is comprehensively detailed in the Table \ref{tab:pretraining_data_detail}. For each data source, we construct temporal sequences using at least one full year of observations per region. This ensures that every sample in our dataset captures complete annual vegetation cycles. Specifically, we define three types of temporal sequences corresponding to the model inputs shown in Figure \ref{fig:flowchart}: Sentinel-2 sequences, Landsat sequences, and MODIS sequences. Due to computational constraints during pre-training, we randomly sample 16 temporally ordered frames from each annual sequence as model inputs.

In Table  \ref{tab:pretraining_data_detail}, "sequence number" refers to the count of temporal sequences using at least one full year of observations per region, while "total number" indicates the aggregate count of individual images across all sequences. The final pre-training dataset comprises 25,244,211 images in total. Notably, we intentionally avoid spatial alignment between different data sources. As illustrated in Figure \ref{fig:flowchart}, each pre-training step simultaneously processes temporally and spatially independent Sentinel-2, Landsat, and MODIS sequences from different locations and timestamps. The model estimates land cover fractions separately for each data source, enabling robust cross-sensor feature learning while maintaining computational efficiency.

\begin{table*}
    \centering
    \small
    \caption{Detailed information of pre-training dataset. "sequence number" refers to the count of temporal sequences using at least one full year of observations per region, while "total number" indicates the aggregate count of individual images across all sequences.}
    \begin{tabular}{c|ccccc}
		\toprule
          Source & Resolution  & Bands & Sequences number & Images number \\
          \midrule
          MODIS & 250m \& 500m  & 7 & 51964 & 1574451 \\
          Landsat-8/9 & 30m  & 6 & 335985 & 13392029 \\
          Sentinel-2 & 10m \& 20m  & 10 & 345843 & 10277731 \\
		 \bottomrule
	\end{tabular}
	\label{tab:pretraining_data_detail}
\end{table*}

\subsection{Method}
{In this section, we introduce our proposed multi-source temporal remote sensing foundation model AgriFM for agriculture mapping}. AgriFM is built based on the modified {Video Swin Transformer which can accept multi-source temporal remote sensing images as input and extract spatio-temporal features from various downsampling stages. Upon the pre-training of the foundation model, a versatile decoder is formulated. This decoder is capable of accepting features extracted from various stages of downsampling in the foundation model. It is specifically designed to generate a spectrum of mapping results, each correlating to its unique set of labels. The overall structure of the framework is shown in Figure \ref{fig:method}.

\begin{figure*}
\centering
\includegraphics[width=\linewidth]{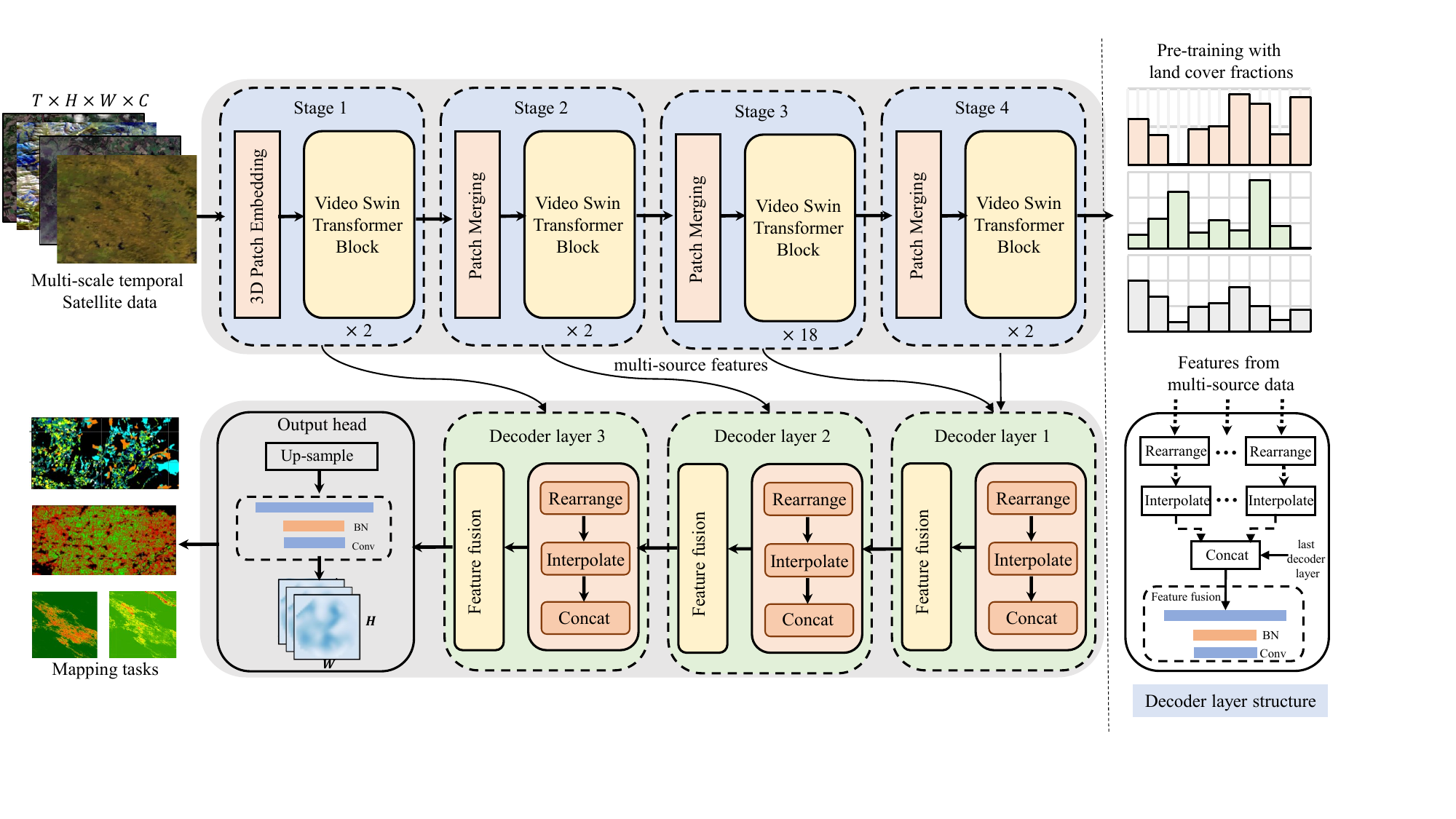}
\caption{Structure of foundation model, AgriFM, comprising four stages. The input satellite sequences (MODIS, Landsat-8/9, and Sentinel-2) are characterized by specific dimensional parameters: $T$ denotes the temporal length of each sequence (randonly selected from 3 to 32 frames), while $W$ and $H$ represent the spatial width and height (both fixed at 224 pixels). The number of spectral bands, $C$, varies depending on the data source. The decoder is purposed for the upsampling and fusion of features to yield mapping results, each marked by their respective labels.}
\label{fig:method}
\end{figure*}

\subsubsection{Multi-source Temporal Foundation Model}

{The input data comprises MODIS data symbolized as $\mathbf{I}_M \in \mathcal{R}^{7\times T\times H \times W}$, Landsat-8/9 data represented as $\mathbf{I}_L \in \mathcal{R}^{6\times T\times H \times W}$, and Sentinel-2 data denoted as $\mathbf{I}_S \in \mathcal{R}^{10\times T\times H \times W}$.}

{To process these multi-source satellite data consistently, we implement separate 3D convolutional patch embedding modules for each modality. Each module employs a single 3D convolutional layer (Conv3D) that simultaneously performs two operations: dividing the input into non-overlapping spatiotemporal patches and projecting them into an embedding space. The kernel size and stride are set identically to the desired patch size $(S_1, D_1, D_1)$, where $S_1$ denotes the temporal patch dimension and $D_1$ the spatial dimension. This configuration ensures comprehensive coverage of the input data cube (Channels × Time × Height × Width) without overlap.}

{For each input patch, the Conv3D operation flattens the pixel values and applies a learned linear projection, outputting a vector of dimension $C_1=128$. The resulting sequences of patch embeddings, which are referred to as MODIS tokens $\mathbf{F}_1^M \in \mathcal{R}^{C_1\times T_1\times H_1 \times W_1}$, Landsat tokens $\mathbf{F}_1^L \in \mathcal{R}^{C_1\times T_1\times H_1 \times W_1}$, and Sentinel tokens $\mathbf{F}_1^S \in \mathcal{R}^{C_1\times T_1\times H_1 \times W_1}$, form the standardized input to the AgriFM backbone. The output dimensions are determined by the patch size parameters as $T_1=\lfloor T/S_1 \rfloor$, $W_1=\lfloor W/D_1 \rfloor$, and $H_1=\lfloor H/D_1 \rfloor$.}

The backbone of AgriFM is constructed based on a modified Video Swin Transformer \citep{liu2022video}. It is proficient at concurrently extracting spatial and temporal features from the input time-series satellite data. The model comprises four hierarchical stages, each containing $2n$ alternating Video Swin Transformer layers: even-numbered layers utilize window-based multi-head self-attention (W-MSA) for local spatiotemporal modeling within non-overlapping $M\times M \times N$ windows, while odd-numbered layers implement shifted window attention (SW-MSA) with shifts along spatial-temporal axes to establish cross-window connections. This dual-attention mechanism effectively balances computational efficiency with global context modeling capabilities. 

{Our foundation model backbone is constructed with four hierarchical Video Swin Transformer stages, each designed to progressively extract and refine spatiotemporal features from multi-source satellite data. The first stage begins with the above patch embedding modules that transform raw satellite imageries from different sources into a unified token representation, establishing a common feature space for subsequent processing. The following three stages (2-4) share a consistent structure, each comprising two key components: a patch merging operation that reduces feature map resolution while increasing channel dimension, followed by a series of transformer blocks that perform spatiotemporal feature learning. These transformer blocks maintain the same architecture as the original Video Swin Transformer, employing shifted window attention mechanisms to efficiently capture both local and global dependencies.}

{This framework is applied consistently across all three satellite data sources (MODIS, Landsat-8/9, and Sentinel-2), with each source processed through identical architectural components. For notational simplicity, we use $\mathbf{F}_i$ and $\mathbf{X}_i$ without superscripts generically to represent the patch merging outputs and transformer block outputs at stage $i$ for any satellite sources. When referring to specific data source, we use superscript notation: $\mathbf{F}_i^M$, $\mathbf{F}_i^L$, $\mathbf{F}_i^S$ and $\mathbf{X}_i^M$, $\mathbf{X}_i^L$, $\mathbf{X}_i^S$ denote the corresponding features for MODIS, Landsat, and Sentinel-2 data respectively.}

{Our architectural modification centers on the synchronized spatiotemporal downsampling in patch merging operation between stages, which fundamentally differs from conventional approaches that only reduce spatial dimensions.  While conventional approaches typically reduce only spatial dimensions, our patch merging module systematically compresses both spatial and temporal resolutions through a coordinated process.}

{For spatial downsampling, we employ a comprehensive sampling strategy: along both height and width dimensions, we initiate sampling from starting positions ranging from 0 to $D_i-1$ with a consistent stride of $D_i$. This method generates exactly $D_i \times D_i$ distinct sampling patterns, each characterized by unique spatial offsets that collectively ensure complete coverage of the input feature maps. For temporal reduction, we implement mean pooling across $S_i$ consecutive frames along the temporal dimension, effectively consolidating temporal information while preserving the essential characteristics of the input sequence.}
{
We then concatenate all resulting features along the channel dimension. This concatenated representation subsequently undergoes a linear projection that expands the channel dimension to $C_{i+1} = 2C_i$, while reducing the spatial dimensions by a factor of $D_i$ in both directions and reducing the temporal dimensions by a factor of $S_i$.
}

{
This operation transforms features from stage $i$, denoted as $\mathbf{F}_i \in \mathbb{R}^{T_i\times H_i\times W_i\times C_i}$, to $\mathbf{X}_i \in \mathbb{R}^{T_{i+1}\times H_{i+1}\times W_{i+1}\times C_{i+1}}$, where the output dimensions are computed as:
}

\begin{equation}
\begin{aligned}
T_{i+1} &= \lfloor T_i/S_i \rfloor, \\
H_{i+1} &= \lfloor H_i/D_i \rfloor, \\
W_{i+1} &= \lfloor W_i/D_i \rfloor, \\
C_{i+1} &= 2C_i.
\end{aligned}
\end{equation}

{
The proposed synchronized spatiotemporal downsampling strategy enables our foundation model to handle dynamically varying input sequence lengths (3-32 frames), making it particularly suitable for agricultural mapping tasks that require flexible temporal modeling across different crop phenological cycles.  In the patch embedding module, we set the temporal patch size to 2 for sequences shorter than 16 frames and to 4 for longer sequences, while maintaining a consistent spatial patch size of 4×4. Subsequent patch merging operations (stages 2-4) uniformly apply a temporal downsampling factor of 2. The inherent property of mean pooling ensures graceful handling of edge cases, when the temporal dimension becomes smaller than the downsampling factor, remaining frames are preserved without information loss. This design enables effective processing of various input sequences (3-32 frames) by progressively learning multi-scale temporal patterns through hierarchical stages, capturing phenological characteristics at varying temporal granularities while maintaining computational efficiency.
}

\subsubsection{Supervised Pre-training with Land Cover Fractions}

{For pre-training the foundation model, we employ land cover fractions as the supervisory signal to guide the learning of semantically meaningful representations. The feature transformation from the final backbone output to land cover fraction prediction involves a carefully designed computational process. The stage 4 output $\mathbf{X}_4 \in \mathbb{R}^{T_4\times H_4\times W_4\times C_4}$ first undergoes spatiotemporal global average pooling to aggregate information across all temporal and spatial positions:}

\begin{equation}
\mathbf{x} = \frac{1}{T_4 \cdot H_4 \cdot W_4} \sum_{t=1}^{T_4} \sum_{h=1}^{H_4} \sum_{w=1}^{W_4} \mathbf{X}_4[:, t, h, w].
\end{equation}
{This operation collapses the spatiotemporal dimensions while preserving the channel-wise information, producing a compact feature vector $\mathbf{x} \in \mathbb{R}^{C_4}$ that encapsulates the global characteristics of the input sequence. The resulting vector subsequently passes through a single-hidden-layer multilayer perceptron (MLP) that performs the final regression:}

\begin{equation}
\hat{\mathbf{p}} = \sigma(\mathbf{W}_2 \cdot \text{ReLU}(\mathbf{W}_1 \mathbf{x} + \mathbf{b}_1) + \mathbf{b}_2),
\end{equation}
{where $\mathbf{W}_1 \in \mathbb{R}^{D \times C_4}$ and $\mathbf{W}_2 \in \mathbb{R}^{K \times D}$ represent the weight matrices of the hidden and output layers respectively, with $D$ denoting the hidden dimension and $K$ the number of land cover categories. The non-linear transformation through ReLU activation enables complex feature interactions, while the final sigmoid function $\sigma$ constrains each element of $\hat{\mathbf{p}}$ to the range [0,1], representing valid fraction estimates. This pre-training framework forces the model to develop representations that directly correspond to physically meaningful land surface properties, establishing a strong foundation for various downstream agricultural monitoring tasks.}

For training objects, we employ the L1 loss function to compute the pre-training losses, utilizing the land cover fractions extracted from GLC\_FCS30D as supervision: 
\begin{equation} 
    L_p=\sum_{i=1}^N |\hat{\mathbf{p}}_i-\mathbf{p}_i|, 
\end{equation} 
where $\hat{\mathbf{p}}_i$ and $\mathbf{p}_i$ correspond to the estimated fractions and the supervised fractions, respectively.

Furthermore, considering the potential for noise interference in the supervised fractions, we implement the mean-teacher method \citep{tarvainen2017mean,li2021geographical} to mitigate the possible impact of noise on foundation model pre-training. We construct a duplicate teacher network, mirroring the structure of the original networks, while referring to the original networks as the student network. During each iteration of pre-training computation, we calculate the fraction as estimated by the teacher network, denoted as $\mathbf{q}$. We then compute the loss function between this teacher-estimated fraction and the student network's output $\hat{\mathbf{p}}$:
\begin{equation}
    L_t=\sum_{i=1}^N |\mathbf{q}_i-\hat{\mathbf{p}}_i|,
\end{equation}
The parameters of the teacher network are not subjected to updates via gradient descent. Rather, they are updated using a moving average method with a minimal step, derived from the student network: 
\begin{equation} 
\theta_{t} = \alpha \cdot \theta_{t} + (1-\alpha) \cdot \theta_{s}, 
\end{equation} 
In this equation, $\theta_{t}$ denotes the teacher network parameters, $\theta_{s}$ represents the student network parameters, and $\alpha$ is a decay factor which governs the pace at which the teacher network parameters are updated.

{The mean-teacher framework establishes a symbiotic learning process where the student network learns from both noisy labels and the teacher's stable predictions, while the teacher gradually integrates the student's knowledge through exponential moving average (EMA) updates. This mutual refinement, regulated by a high smoothing coefficient ($\alpha$), ensures consistent convergence direction. As pre-training progresses, the teacher's parameters evolve into a robust temporal ensemble of the student's weights, effectively filtering label noise while maintaining consensus between both models.}

\subsubsection{Versatile Decoder for Unified Mapping}\label{sec:decoder}

The multi-source temporal foundation model enables the unified crop  mapping to process heterogeneous temporal remote sensing data from multiple sources, facilitating feature extraction at different sampling stages. While the foundation model provides powerful feature extraction capabilities, it cannot directly generate mapping results. {To effectively utilize these multi-scale temporal features, we design a versatile decoder architecture that unifies various agricultrue mapping tasks. }

For a given mapping task with $K$ input data sources (where $K$ represents any combination of MODIS, Landsat-8/9, and Sentinel-2 data sources), the foundation model first extracts features from each source independently. 
The decoder reconstructs a single crop map through three upsampling layers, each doubling resolution via upsampling and convolutional layers. Decoder layer $j$ fuses two inputs: 1) upsampled features $U_{j-1}$ from the previous layer (initialized with stage 4 encoder outputs $X_4$), and 2) skip-connected features $X_i$ from encoder stage $i=4-j$. Each decoder layer contains four key operations: rearrange, interpolate, concat, and feature fusion.  Given features $\mathbf{F}_{i} \in \mathbb{R}^{T_i \times H_i \times W_i \times C_i}$ from the $i$-th stage of the foundation model, the rearrange operation combines the temporal and channel dimensions to produce $\mathbf{F}_{i}^{'} \in \mathbb{R}^{ H_i \times W_i \times (T_i \times C_i)}$. The interpolate function then performs upsampling to match the required spatial dimensions. The concat operation merges the upsampled features with the previous decoder layer's output along the channel dimension, yielding $\mathbf{U}^c_j \in \mathbb{R}^{H_i \times W_i \times (K \times C_i + C_{i-1})}$. Finally, the feature fusion module, implemented with two convolutional layers and a batch normalization layer, processes the concatenated features to generate the decoder layer output $\mathbf{U}_{j}$.

For tasks requiring auxiliary data inputs, our framework can directly integrate additional feature maps (e.g., from CNNs) by bypassing the rearrange operation. The final decoder output is processed through a classification layer to produce mapping results. The training objective minimizes the cross-entropy loss:
\begin{equation}
L=-\sum_{i} y_{i} \log(\hat{y}_{i})
\end{equation}
where $y_{i}$ denotes the ground truth label and $\hat{y}_{i}$ represents the predicted probability. To address the prevalent class imbalance in agriculture mapping tasks, we implement a hard sample mining strategy that focuses learning on challenging examples during training.

\subsection{Downstream Mapping Tasks and Validation}
{Since our goal is to construct the foundation model for agriculture  mapping, our validation emphasis lies in ensuring the optimal performance of our method when labels are available. On one hand, we seek to validate the efficacy of our proposed AgriFM in handling a variety of agriculture mapping tasks, while on the other hand, we aim to demonstrate that it outperforms the existing remote sensing foundation models. To this end,  we have selected five representative tasks across three regions for validation, including  1) agricultural land mapping, 2) field boundary delineation 3) fine-grained agricultural land use/land cover mapping, 4) paddy field mapping, and 5) winter wheat mapping.} The satellite data and other detailed information required for validating these tasks are shown in Table \ref{tab:agriculture_data_detail}. {The table provides detailed information on all downstream tasks used for evaluating AgriFM. For each task, we specify the geographic region, data split (training/validation/testing), acquisition year, satellite sources, spatial resolution, spectral bands, temporal length (number of observations per sample), number of images, and spatial size (pixel dimensions). The temporal length indicates the input sequence size for each task. The images column represents the total sample count for each split, while size denotes the spatial dimensions of input patches in pixels.
} The spatial distribution of these data is illustrated in Figure \ref{fig:dist_agriculture}.

\subsubsection{{Agricultural Land Mapping, Field Boundary Delineation and Agricultural Land Use/Land Cover Mapping}}

{For the tasks of agricultural land mapping, field boundary delineation and agricultural land use/land cover mapping in the Auvergne-Rhône-Alpes (ARA) region of France, we derive labels from the EuroCrops dataset \mbox{\protect \citep{schneider2023eurocrops}}. EuroCrops is an all-encompassing dataset featuring geo-referenced polygons of agricultural croplands that cover 16 countries within the European Union (EU). It also provides data on specific crop types cultivated across these regions. This dataset allows us to obtain a comprehensive distribution of crops in the region from 2018 to 2020. Using this information, we generate labels for both agriculture land areas and boundaries, and select  16  agriculture land use/land cover as labels for the agricultural land use/land cover mapping task.}  We utilize Sentinel-2 data from three consecutive years, 2018, 2019, and 2020, for training, validation, and testing respectively. Similar to the pre-training data processing, we aggregate the 10-meter and 20-meter resolution bands of the Sentinel-2 satellite into a unified 10-meter resolution, resulting in data across 10 spectral bands. For each task, we randomly select 2327 images, each measuring $256\times 256$ pixels.

{For agricultural land mapping and field boundary delineation, we utilize the first 32 available scenes from the first six months of each year as model inputs. In contrast, the agricultural land use/land cover mapping task employs a distinct temporal sampling strategy: we select 4 scenes per month to construct a 24-scene time series.} This design serves dual purposes: (1) to explicitly validate our model's capability in handling variable-length temporal inputs, and (2) to incorporate prior knowledge from previous year's cropping patterns. The prior information, extracted using a ResNet50 \mbox{\protect \citep{he2016deep}} backbone, is concatenated with the foundation model's output features. To maintain computational efficiency while ensuring feature richness, we deliberately limit the input sequence to 24 frames for this specific task.
These experiments are designed to assess the effectiveness of our method in addressing high-resolution mapping tasks over extended temporal scales.

\begin{figure*}[!htb]
\centering
\includegraphics[width=0.9\linewidth]{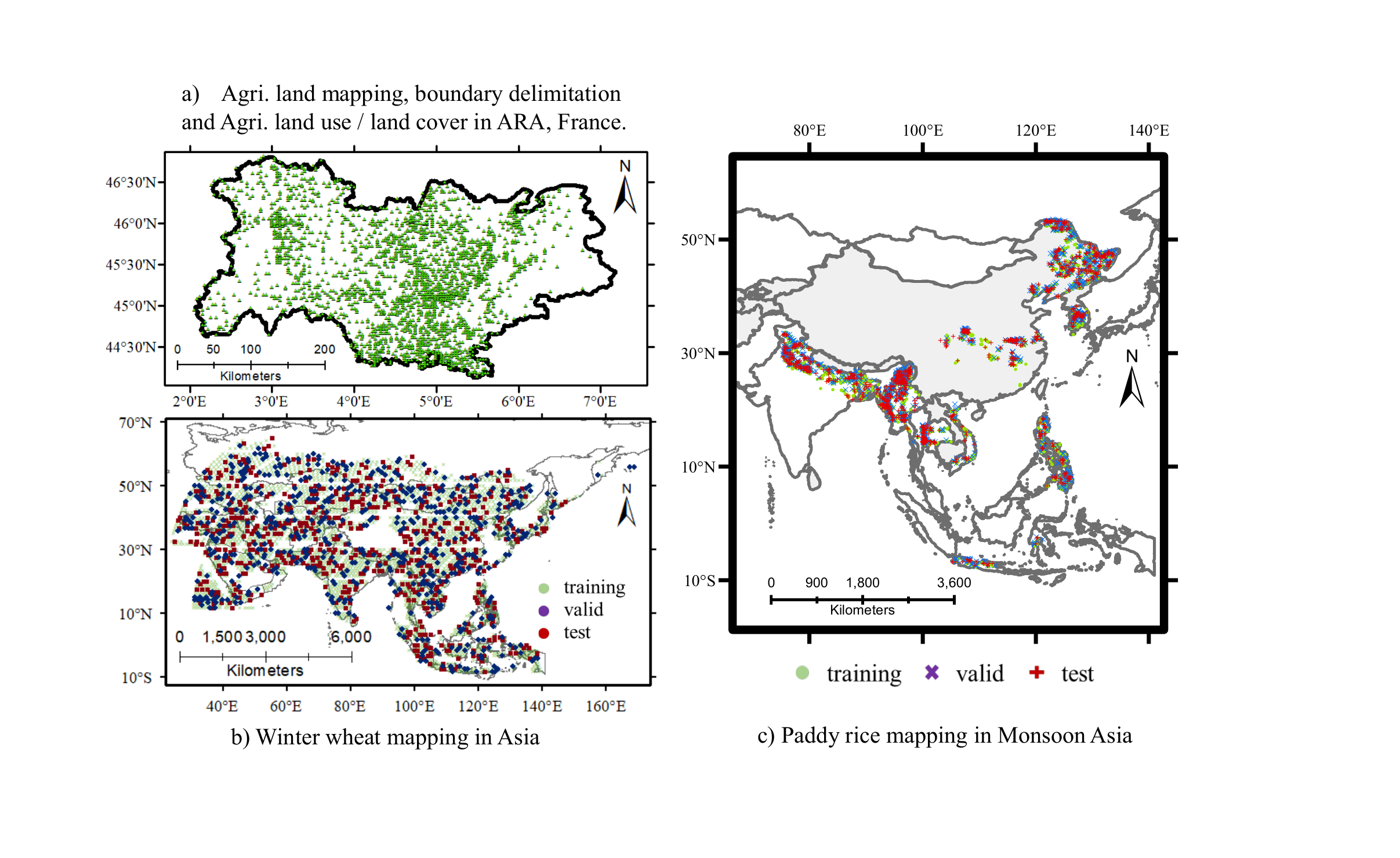}
\caption{{The study area and dataset detailed information for downstream mapping tasks.}}
\label{fig:dist_agriculture}
\end{figure*}

\begin{table*}
    \centering
    \small
    \caption{{Summary of downstream agricultural mapping tasks and dataset specifications. The table details the geographic region, data split (training/validation/testing), acquisition year, satellite sources, spatial resolution, spectral bands, temporal sequence length, number of sample images, and spatial dimensions for each task. Temporal indicates the number of input frames per sample; Images refers to the total count of labeled instances; Size denotes the spatial dimensions in pixels.}}
    \resizebox{\linewidth}{!}{
    \begin{tabular}{c|ccccccccc}
		\toprule
          Tasks & Region & Split & Year & Sources & Resolution & Bands & Temporal & Images & Size \\
          \midrule
          \multirow{3}{*}{\makecell{Agricultural Land Mapping \\ Field Boundary Delineation}} & ARA, France& training & 2018 & Sentinel-2 & 10m & 10 & 32 & 2327 & 256\\
          & ARA, France & validation & 2019 & Sentinel-2 & 10m & 10 & 32 & 2327  & 256\\
          & ARA, France & testing & 2020 & Sentinel-2 & 10m & 10 & 32 & 2327  & 256\\
          \midrule
          \multirow{3}{*}{\makecell{Agricultural Land Use\\/Land Cover Mapping}} & ARA, France& training & 2018 & Sentinel-2 & 10m & 10 & 24 & 2327 & 256\\
          & ARA, France & validation & 2019 & Sentinel-2 & 10m & 10 & 24 & 2327  & 256\\
          & ARA, France & testing & 2020 & Sentinel-2 & 10m & 10 & 24 & 2327  & 256\\
          \midrule
          \multirow{3}{*}{Paddy Rice Mapping} & Monsoon Asia & training & 2019 & HLS30 & 30m  & 6 & 5 & 3039 &224 \\
          & Monsoon Asia& validation & 2019 & HLS30 & 30m & 6 & 5 & 1013 &224 \\
          & Monsoon Asia & testing & 2019 & HLS30 & 30m  & 6 & 5 & 1013 &224 \\
          \midrule
          \multirow{3}{*}{Winter Wheat Mapping} & Asia & training & 2021 &  MODIS & 250m & 7 & 11 & 2711 & 512 \\
          & Asia & validation & 2021 &MODIS & 250m & 7 & 11 & 636 & 512 \\
          & Asia & testing & 2021 & MODIS & 250m & 7 & 11 & 569 & 512 \\
		 \bottomrule
	\end{tabular}
        }
	\label{tab:agriculture_data_detail}
\end{table*}

\subsubsection{{Paddy Rice Mapping in Monsoon Asia}}
For the task of rice mapping, we adopt the method proposed by Fang et al. \citep{fang5138538generating} to generate labels from areas of high confidence by comparing and integrating existing products. We select five HLS30 (Harmonized Landsat and Sentinel-2) images from the first six months of 2019, prioritizing those with minimal cloud coverage, to serve as input data. Given that rice is typically cultivated in regions with high cloud coverage, the inclusion of additional temporal data is generally advantageous. Thus, we enrich our dataset with MODIS data, which boasts a higher temporal resolution. Both data sources, despite their differing resolutions, are effectively harnessed for rice mapping. This task is designed to evaluate the efficacy of our method in rice mapping and demonstrate how our proposed method can effectively leverage the benefits of integrating multiple data sources.

Specifically, the rice mapping tasks are conducted in the Monsoon Asia region, utilizing 2019 data for training, validation, and testing. The input data are sourced from both HLS30 and MODIS, with respective resolutions of 30m and 250m, and providing 6 and 7 spectral bands. The sequence length for the HLS30 and MODIS data is 5 and 46, respectively. In total, 3039 images are used for training, while 1013 images are used for both validation and testing. The image sizes for the HLS30 and MODIS data are 224 and 56, respectively.

\subsubsection{{Winter Wheat Mapping  in Asia}}
For the task of winter wheat mapping, we follow the method proposed by Li et al. \citep{li5029097asiawheat} to generate labels for winter wheat in Asia based on the WorldCereal product \citep{van2023worldcereal}. In accordance with the growth cycle of winter wheat in Asia, we select one MODIS image per month from August 2020 to June 2021 to form the training input sequence.
Specifically, the winter wheat mapping tasks are carried out in Asia, using MODIS data for training, validation, and testing. The input data are derived from MODIS with a resolution of 250m, providing 7 spectral bands. The sequence length for the MODIS data is 11. In total, 2711 images are used for training, 636 images for validation, and 569 images for testing. The image size for the MODIS data is 512.

\subsubsection{{Multi-source Mapping Tasks}}
{To evaluate the cross-spatial generalization capabilities of AgriFM, we construct a multi-source dataset incorporating Sentinel-2, Landsat 8/9, and MODIS satellite imagery with varying temporal configurations. For Sentinel-2, we utilize 32 temporal observations for agricultural land mapping and field boundary delineation, and 24 observations for agriculture land use/cover mapping. Landsat 8/9 data contained 12 frames, while MODIS provided 44 temporal observations. This configuration enables systematic analysis of model performance across different spatial resolutions and temporal densities.}

{The dataset design specifically addresses cross-spatial generalization by including both single-source and multi-source combinations. We evaluate pure Sentinel-2 scenarios against Landsat-only, Landsat-MODIS, Landsat-Sentinel fusion setups, allowing direct comparison of how different spatial and temporal characteristics affect mapping accuracy across all agricultural tasks. This structured approach provides comprehensive insights into the model's ability to leverage complementary information from multi-resolution satellite systems.}

\subsubsection{{Validation}}
We use precision (P), recall (R), F1 score (F1), and Overall Accuracy (OA) as our validation metrics:

\begin{equation}\centering
\begin{split}
      P = &\frac{TP}{TP + FP},\\
      R = &\frac{TP}{TP + FN},\\
      F1 =& 2 \cdot \frac{\text{P} \cdot \text{R}}{\text{P} + \text{R}}\\
     OA = &\frac{TP + TN}{TP + FP + TN + FN},
\end{split}
\end{equation}
where $TP, FP, TN, FN $ represent the number of true positive samples, false positive samples, true negative samples and false negative samples.

To comprehensively evaluate AgriFM's performance, we conduct comparisons across three distinct methodological categories. First, we examine conventional deep learning architectures widely adopted in various mapping tasks: (1) A CNN (VGG16 variant \citep{simonyan2014very}) treating temporal observations as additional input channels; (2) A CNN-LSTM hybrid extracting spatial features per timestep using VGG16 followed by temporal modeling through LSTM; and (3) A 3D CNN processing spatiotemporal cubes (height × width × time × bands). All baselines employ identical decoder structures and training protocols to isolate architectural differences, ensuring fair comparison of their spatiotemporal representation capabilities. { While more advanced CNN variants (e.g., ResNet50 \citep{he2016deep}) exist, VGG16 provides a balanced representation of CNN capabilities through its straightforward yet effective hierarchical feature extraction design. This choice allows us to focus on comparing fundamental architectural paradigms, CNNs versus transformer-based foundation models, rather than optimizing performance through model-specific refinements.}

Second, we compare against ViT-based foundation models that incorporate temporal information during pretraining. Specifically, we evaluate SatMAE \citep{cong2022satmae} (using Landsat-8 data) and Prithvi \citep{Prithvi-100M-preprint} (trained on HLS30 data), which represent the most temporally-aware open-source RSFMs currently available, despite their limited sequence lengths. {In addition, we also incorporate some recently released RSFMS, Galileo \citep{tseng2025galileo} and SMARTIES \citep{sumbul2025smarties} to verify our method's effeveness.} These comparisons reveal how feature fusion strategies and hierarchical representations impact mapping precision, particularly for crops with subtle spectral-temporal signatures.

Third, we analyze PIS \citep{PIS} and {GFM \citep{GFM}, two Swin Transformer-based foundation models lacking explicit temporal processing.} While sharing similar backbone architecture and pretraining data scales with our approach, their non-temporal designs help isolate the benefits of our temporal-aware pretraining strategy. Although they use the Swin transformer as its backbone, we can reuse some weights as pre-trained weights of the video Swin transformer and adopt the same architecture as our method for comparison.

\subsubsection{Implementation details}
We develop our codes using Python 3.11 and PyTorch-2.0. During the pre-training phase, the network learning rate is set to 1e-5. To ensure the stability of pre-training, the learning rate gradually increases from 1e-7 to 1e-5 over the first 5000 iterations, after which it gradually decreases throughout the pre-training process until it reaches 1e-6. The pre-training phase lasts for 50 epochs. The training is carried out on ten L40 GPUs, with a batch size of 80. The coefficient for the moving average is set to 0.001.

For downstream mapping tasks, we employ the same network structure, that is, the unified mapping framework that we construct. The learning rate is set to 6e-5. All experiments are conducted on four L40 GPUs, with a batch size of 16, and last for 50 epochs. We select the model that performs best on the validation dataset as our final result.

\section{{Agricultural Land Mapping and Field Boundary Delineation}}

This experiment evaluates the AgriFM's performance for cropland mapping and field boundary delineation with time series Sentinel-2 images. Utilizing three consecutive years (2018-2020) of Sentinel-2 data over France's Auvergne-Rhône-Alpes (ARA) region, we process 10 spectral bands aggregated to 10m resolution, with 32 temporal steps per annual sequence. The dataset comprises 2,327 geo-referenced tiles (256×256 pixels) per year, strictly partitioned by temporal years: 2018 for training, 2019 for validation, and 2020 for testing. Boundary delineation presents distinct challenges from cropland mapping, requiring precise field edge detection that demands higher spatial granularity \citep{d2023ai4boundaries, persello2023ai4smallfarms}.

As detailed in Tables \ref{tab:cropland_metric} and \ref{tab:boundary_metric}, {we systematically evaluate three methodological categories: (1) CNN variants (CNN, CNN-LSTM, 3DCNN) representing conventional deep learning approaches; (2) ViT-based methods (SatMAE, Prithvi, Galileo and SMARTIES) as current mainstream remote sensing foundation models pretrained with temporal data; and (3) Swin transformer-based methods (PIS, GFM, AgriFM-scratch) demonstrating hierarchical feature extraction's impact on agriculture mapping. Notably, backbones of PIS and GFM are Swin transformers and don't adopt temporal data. We adaptively rearrange its partial parameters to initialize our Video Swin Transformer, enabling comparison of temporal-agnostic pretraining effects. AgriFM-scratch serves as the randomly initialized control. The "Positive" column specifies agricultural land / boundary-specific metrics, while "Average" incorporates background class performance.} Figure \ref{fig:cropland_comparison} visually contrasts core metrics (Precision / Recall / F1) across methods, and Figure \ref{fig:cropland_mapping_results} showcases regional mapping outputs where green area denotes cropland and red area indicates boundaries.

\begin{table}[!htb]
\caption{{Performance comparison on the agricultural land mapping task (\%). The positive column means metrics of agricultural land identification, while the average columns means the average metric of both background and agricultural land.}}
\label{tab:cropland_metric}
\centering
\small
\begin{tabular}{l|ccccccc}
\toprule
\multirow{2}{*}{\textbf{Model}} & \multicolumn{3}{c}{\textbf{Positive}} & \multicolumn{4}{c}{\textbf{Average}}  \\
 \cmidrule(lr){2-4} \cmidrule(lr){5-8} 
 & Precision & Recall & F1 & Precision & Recall & F1 & OA \\

\midrule
CNN & 73.30 & 71.71 & 73.00 & 75.03 & 75.00 & 75.01 & 75.18 \\
CNN-LSTM & 79.74 & 81.06 & 80.39 & 81.63 & 81.70 & 81.66 & 81.75 \\
3DCNN & 80.30 & 75.70 & 77.93 & 80.22 & 79.89 & 80.00 & 80.21 \\
Prithvi (ViT-b) & 74.66 & 72.73 & 73.69 & 75.90 & 75.79 & 75.83 & 76.02 \\
Galileo (ViT-b) & 76.71 & 74.78 & 75.73 & 77.81 & 77.70 & 77.74 & 77.93 \\
SatMAE (ViT-L) & 76.85 & 76.76 & 76.80 & 78.47 & 78.47 & 78.47 & 78.60 \\
SMARTIES (ViT-L) &  75.04 & 74.48 & 74.76 & 76.70 & 76.67 & 76.68 & 76.84\\
PIS (Swin-b) & 80.11 & 77.94 & 79.01 & 80.81 & 80.67 & 80.73 & 80.88 \\
GFM (Swin-b) & 80.99 & 81.12 & 81.06 & 82.39 & 82.40 & 82.39 & 82.50\\
AgriFM-scratch & 79.53 & 78.47 & 79.00 & 80.64 & 80.58 & 80.60 & 80.74 \\
AgriFM & \textbf{84.31} & \textbf{81.90} & \textbf{83.09} & \textbf{84.58} & \textbf{84.42} & \textbf{84.48} & \textbf{84.61} \\
\bottomrule
\end{tabular}
\end{table}

\begin{table}[!htb]
\caption{{Performance comparison on the field boundary delineation task (\%). The positive column means metrics of field boundary identification, while the average columns means the average metric of both background and field boundary.}}
\label{tab:boundary_metric}
\centering
\small 
\begin{tabular}{l|ccccccc}
\toprule
\multirow{2}{*}{\textbf{Model}} & \multicolumn{3}{c}{\textbf{Positive}} & \multicolumn{4}{c}{\textbf{Average}}  \\
 \cmidrule(lr){2-4} \cmidrule(lr){5-8} 
 & Precision & Recall & F1 & Precision & Recall & F1 & OA \\
\midrule
CNN & 67.20 & 67.90 & 67.55 & 73.84 & 73.93 & 73.88 & 75.42 \\
CNN-LSTM & 73.53 & 70.92 & 72.20 & 78.16 & 77.74 & 77.94 & 79.43 \\
3DCNN & 68.85 & 66.98 & 67.90 & 74.61 & 74.33 & 74.46 & 76.14 \\
Prithvi (ViT-b) & 52.81 & 54.10 & 53.45 & 62.32 & 62.44 & 62.38 & 64.50 \\
Galileo (ViT-b) &  59.86 & 65.41 & 62.51 & 68.91 & 69.51 & 69.11 & 70.52\\
SatMAE (ViT-L)  & 59.13 & 66.28 & 62.50 & 68.57 & 69.29 & 68.78 & 70.04 \\
SMARTIES  (ViT-L)   &  57.30 & 62.78 & 59.92 & 66.77 & 67.31 & 66.94 & 68.44 \\
PIS (Swin-b) & 70.44 & 71.98 & 71.20 & 76.64 & 76.86 & 76.70 & 78.07 \\
GFM (Swin-b) & 74.54 & 70.33 & 72.38 & 78.60 & 77.91 & 78.21 & 79.78\\
AgriFM-scratch & 70.63 & 73.70 & 72.13 & 77.15 & 77.59 & 77.35 & 78.55 \\
AgriFM & \textbf{75.11} & \textbf{77.47} & \textbf{76.27} & \textbf{80.61} & \textbf{80.98} & \textbf{80.78} & \textbf{81.84} \\
\bottomrule
\end{tabular}
\end{table}

For agricultural land  mapping (Table \ref{tab:cropland_metric}), CNN architectures demonstrate strong baseline performance, with CNN-LSTM leading at 80.39\% F1-score due to its explicit spatiotemporal fusion design. 3DCNN achieves only 77.93\% F1 despite 3D convolutions. This shows that simply adding 3D convolution does not guarantee better spatiotemporal modeling, and also shows that the 3D convolution in traditional deep learning methods for processing spatiotemporal data may have insufficient information extraction capabilities for mapping tasks. {ViT-based foundation models trail slightly at 73.69\%, 75.73\%, 76.80\% and 74.76\% respectively, confirming ViT's architectural constraints. The 16×16 patch embedding causes irreversible spatial detail loss} (evident in Figure \ref{fig:cropland_mapping_results}'s blurred boundaries). While increasing input resolution could theoretically mitigate this issue, the computational cost becomes prohibitive for temporal sequences while distorting original resolution relationships. {The Swin-based PIS and GFM achieves 79.01\% and 81.06\% F1, outperforming all ViT variants and underscoring hierarchical features' importance, despite its non-temporal pretraining. AgriFM dominates with 83.09\% F1, affirming both Video Swin's architectural superiority for agriculture mapping and the added value of temporal-aware pretraining.} 

\begin{figure*}
\centering
\includegraphics[width=\linewidth]{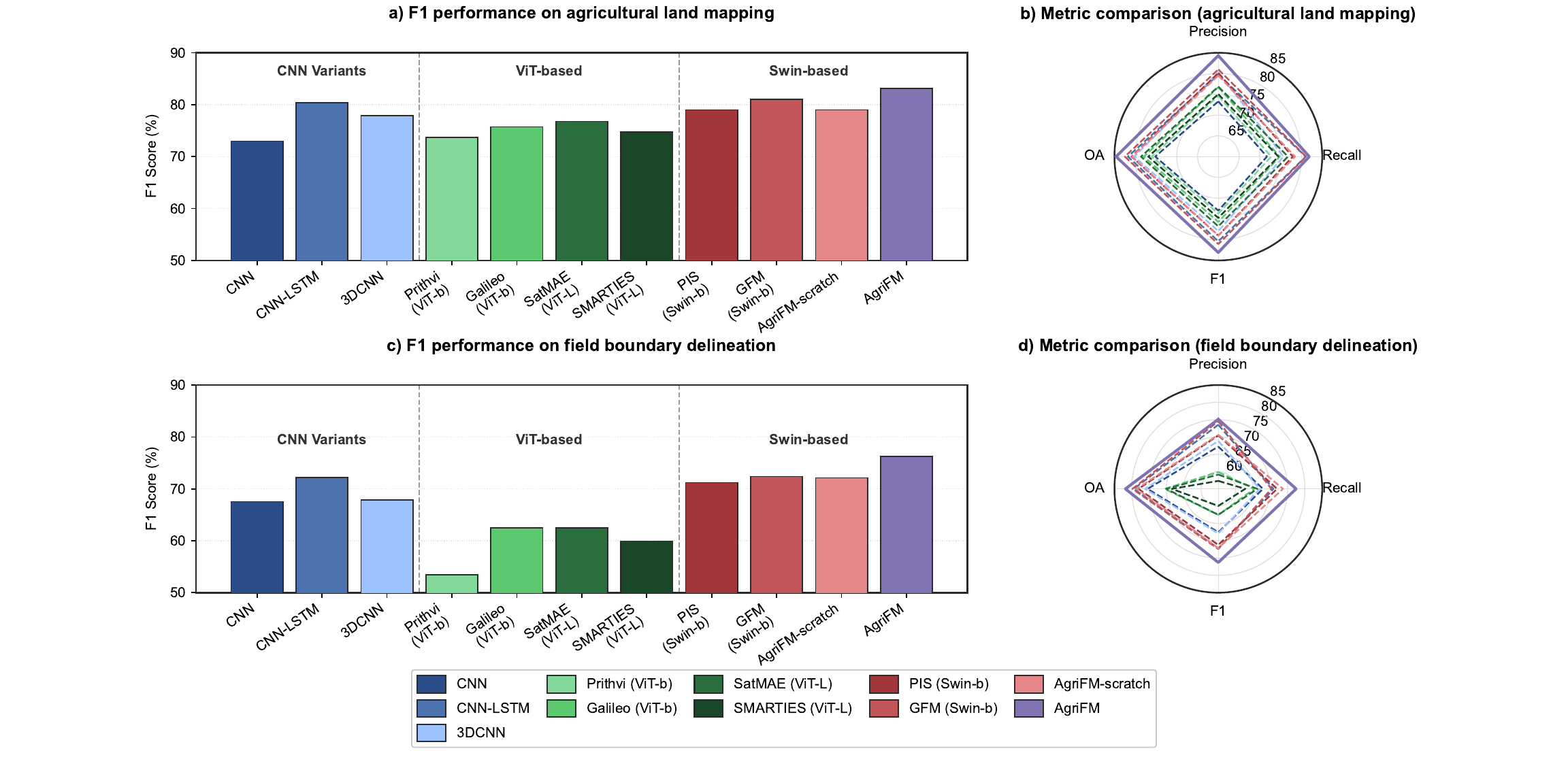}
\caption{{Comparative performance on agricultural land mapping and boundary delineation task. Bar plots show F1-score for the positive class across models, grouped by architecture type (CNN variants, ViT-based, and  Swin-based). Radar plots show metrics (precision, recall, F1-score and OA) comparison across models.}}
\label{fig:cropland_comparison}
\end{figure*}

Boundary delineation (Table \ref{tab:boundary_metric}) further accentuates methodological differences. {ViT-based methods performance plummets (Prithvi: 53.45\% F1; Galileo: 62.51\%; SatMAE: 62.50\% F1; SMARTIES: 59.92\% F1),} consistent with Figure \ref{fig:cropland_vis_results}'s oversmoothed edges—ViT's aggressive downsampling struggles with sub-pixel boundary variations. CNN-LSTM maintains relative robustness (72.20\% F1). {PIS and GFM show architectural promise (71.20\% F1 and 72.38\% F1) but still trail AgriFM by 5.07\% and 3.89\%, primarily due to lacking temporal pretraining. Intriguingly, randomly initialized AgriFM-scratch achieves 72.13\% F1, merely 4.14\% below AgriFM, suggesting hierarchical backbone inherently excels at spatial sensitivity while temporal pretraining provides systematic refinement. The underperformance of PIS and GFM versus Agri-scratch highlights incompatibilities when rearranging non-temporal pretrained weights, justifying our dedicated spatiotemporal pretraining strategy.}

\begin{figure*}
\centering
\includegraphics[width=0.8\linewidth]{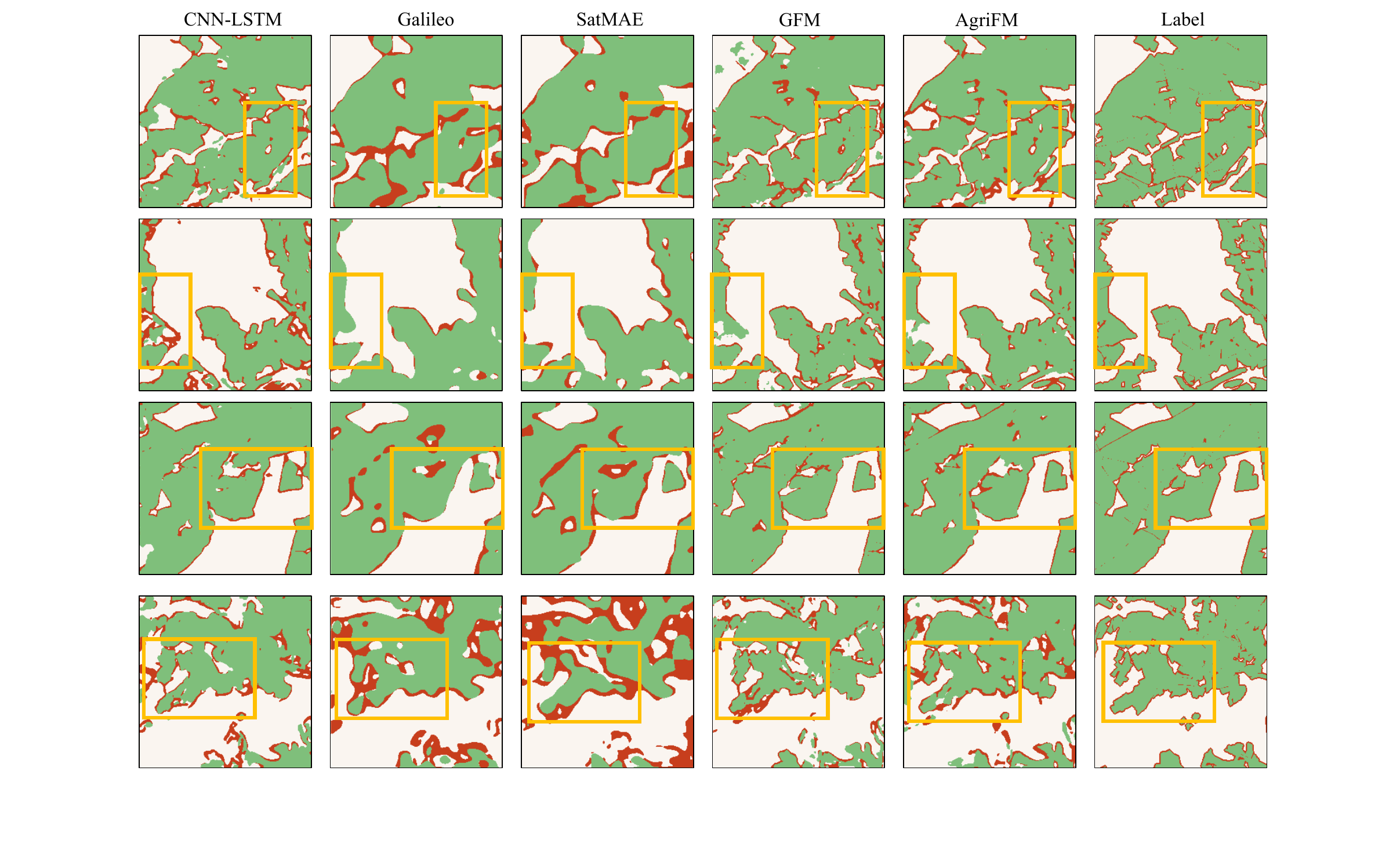}
\caption{{Visual comparison of agricultural land mapping and field boundary delineation results from test dataset in 2020.  Only  the best-performing methods from each task are shown to ensure clarity in comparison. Cropland pixels are shown in green and field boundaries in red. Representative regions with notable differences are highlighted with yellow bounding boxes for detailed comparison.}}
\label{fig:cropland_vis_results}
\end{figure*}

Three principal conclusions emerge from above experiments and analysis: (1) ViT's aggressive downsampling fundamentally limits agricultural applications requiring fine spatial outputs; (2) Video Swin transformer's hierarchical design shows superior potential when combined with temporal pretraining; (3) AgriFM synergizes architectural and pretraining innovations to deliver both global spatiotemporal understanding and pixel-level precision.

We utilize our trained AgriFM to generate a map of cropland and its boundaries in the ARA region of France for the year 2020, as illustrated in Figure \ref{fig:cropland_mapping_results}. The green areas represent cropland, while the red areas delineate the boundaries of this cropland. 

\begin{figure*}
\centering
\includegraphics[width=0.8\linewidth]{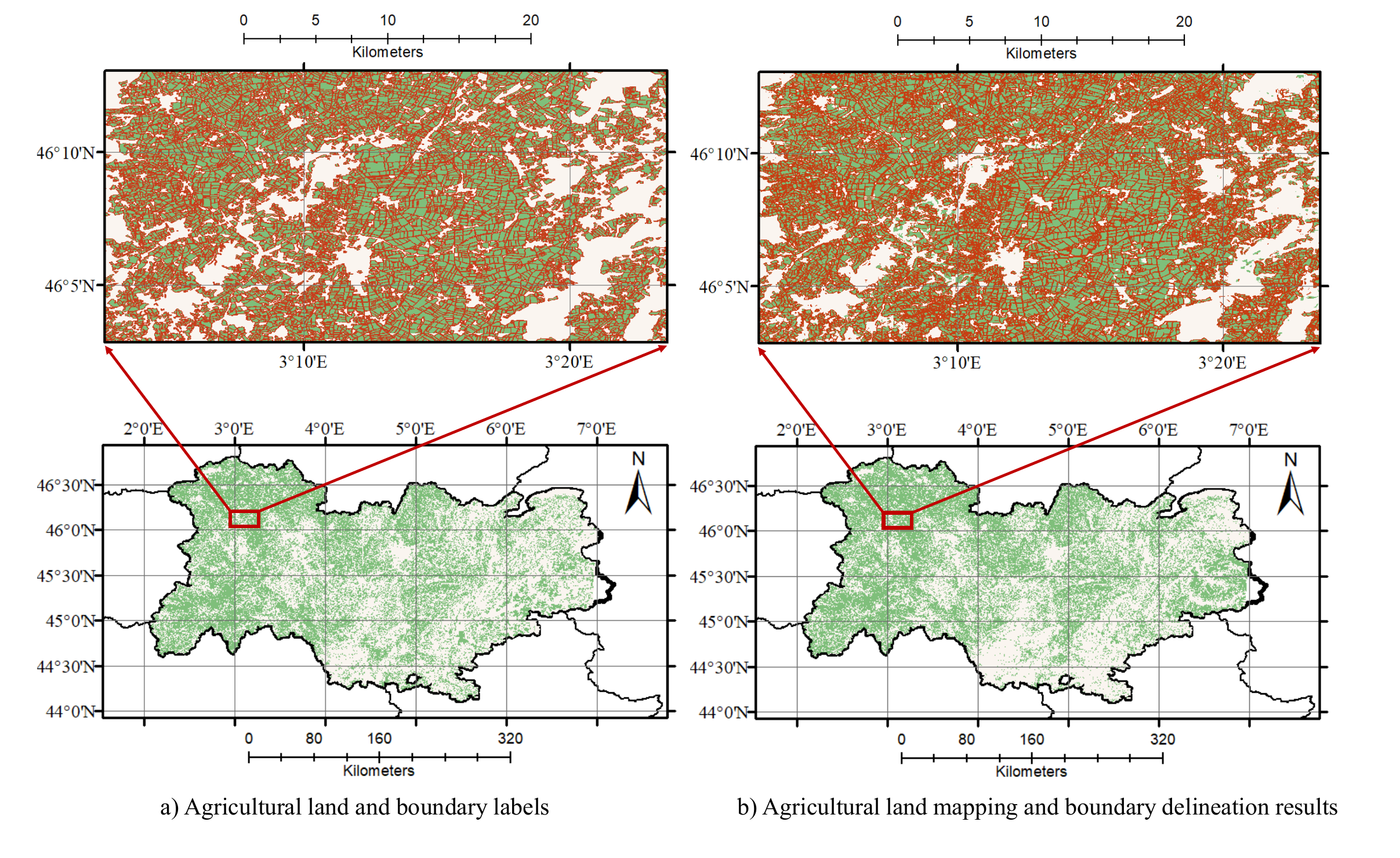}
\caption{{The generated map of agricultural land and its boundaries in the ARA region of France for the year 2020: a) labels from EuroCrops dataset; b) the mapping results from our method. The lower panel shows the complete mapping coverage across the study area, while the upper panel provides enlarged views of selected representative regions. Green areas represent identified cropland, with red boundaries delineating individual field parcels. }}
\label{fig:cropland_mapping_results}
\end{figure*}

As indicated in the generated maps, our proposed model AgriFM demonstrates high effectiveness in mapping cropland and delineating its boundaries. However, it is worth noting that in areas where cropland is less dense, instances of under-detection may occur. Furthermore, while the method excels at defining boundaries within the cropland, the discontinuity at the borders of the cropland presents a challenge. This is an inherent limitation of the approach we have employed, as the paradigm rooted in semantic segmentation cannot ensure the continuity of the segmentation results.

\section{{Agricultural Land Use / Land Cover Mapping}}
This experiment specifically assesses AgriFM's capability to recognize agricultural land use / land cover during early growth stages—a critical requirement for proactive agricultural management. Leveraging the same Sentinel-2 dataset (10m resolution, 10 spectral bands) and spatial coverage (ARA region, France) as the agricultural land mapping task,  we specifically target classification of 16 major classes during their initial six-month growth phases using 2018-2020 data. The temporal sequence construction adopts a stratified sampling approach: four observations per month are selected, generating 24 temporal observations  that capture early phenological development. The 2,327 tiles per year maintain strict temporal isolation: 2018 (training), 2019 (validation), and 2020 (testing). 

As shown in Table \ref{tab:early_crops_metric}, we systematically compare various methods with four metrics: precision, recall, F1-score, and overall accuracy (OA). Figure \ref{fig:early_comparison} presents three complementary visualizations: (a) bar charts comparing F1 scores, (b) radar plots showing metric balance, and (c) line graphs analyzing F1-scores across land cover / land use ordered by training sample frequency (low to high). Figure \ref{fig:early_results} further displays spatial mapping comparisons in representative regions.

\begin{table}[!htb]
\caption{{Performance comparison on agricultural land use / land cover mapping (\%).  Metrics are computed as average of each class. }}
\label{tab:early_crops_metric}
\centering
\small 
\begin{tabular}{l|cccc}
\toprule

Model & Precision & Recall & F1-score &  OA \\
\midrule
CNN            & 43.50  & 36.71  & 40.12  & 73.04  \\
CNN-LSTM       & 52.47  & 40.23  & 44.29  & 74.74  \\
3DCNN          & 46.57  & 36.81  & 40.60  & 73.71  \\
Prithvi  (ViT-b)  & 60.00  & 37.16  & 42.10  & 68.68  \\
Galileo  (ViT-b)  & 60.20 & 39.28 & 45.12 & 70.41  \\
SatMAE (ViT-L)    & 64.80  & 40.24  & 46.10  & 71.03  \\
SMARTIES  (ViT-L)  &  60.33 & 41.51 & 47.65 & 71.57\\
PIS (Swin-b)  & 67.11  & 48.63  & 54.51  & 76.18  \\
GFM (Swin-b) & \textbf{70.55} & 51.08 & 57.75 & 76.99\\
Swin-scratch   & 70.43  & 48.35  & 54.82  & 76.26  \\
AgriFM         &  68.34 & \textbf{54.97} & \textbf{60.49} & \textbf{77.38} \\
\bottomrule
\end{tabular}
\end{table}

{The results demonstrate AgriFM's clear superiority with 60.49\% F1 and 77.38\% OA. Architecturally similar approaches (PIS: 54.51\% F1; GFM: 57.75\%; AgriFM-scratch: 54.82\% F1) also significantly outperform alternatives. Notably, the performance gaps between CNN / ViT methods and AgriFM (13-15\% F1) are markedly larger than in the agricultural land mapping task, underscoring the heightened challenge of agricultural land use / land cover mapping requiring precise spatiotemporal feature coordination. Three key observations emerge: (1) CNN variants (best: CNN-LSTM 44.29\% F1) excel at spatial feature fusion but lack temporal modeling capacity; (2) ViT-based methods (best: SMARTIES 47.65\% F1) leverage global attention for temporal modeling but sacrifice spatial detail; (3) AgriFM uniquely integrates both capabilities through multi-source temporal modeling and pretraining.}

Figure \ref{fig:early_comparison} c) reveals distinct response patterns to class imbalance. {ViT-based methods show advantages for rare classes (left side), benefiting from global temporal context capture. Conversely, CNNs outperform ViTs  for dominant classes (right side), where local spatial features prove more discriminative. AgriFM maintains remarkably stable performance across all frequency categories, achieving high F1 improvement for rare classes  versus suboptimal methods.}

\begin{figure*}
\centering
\includegraphics[width=\linewidth]{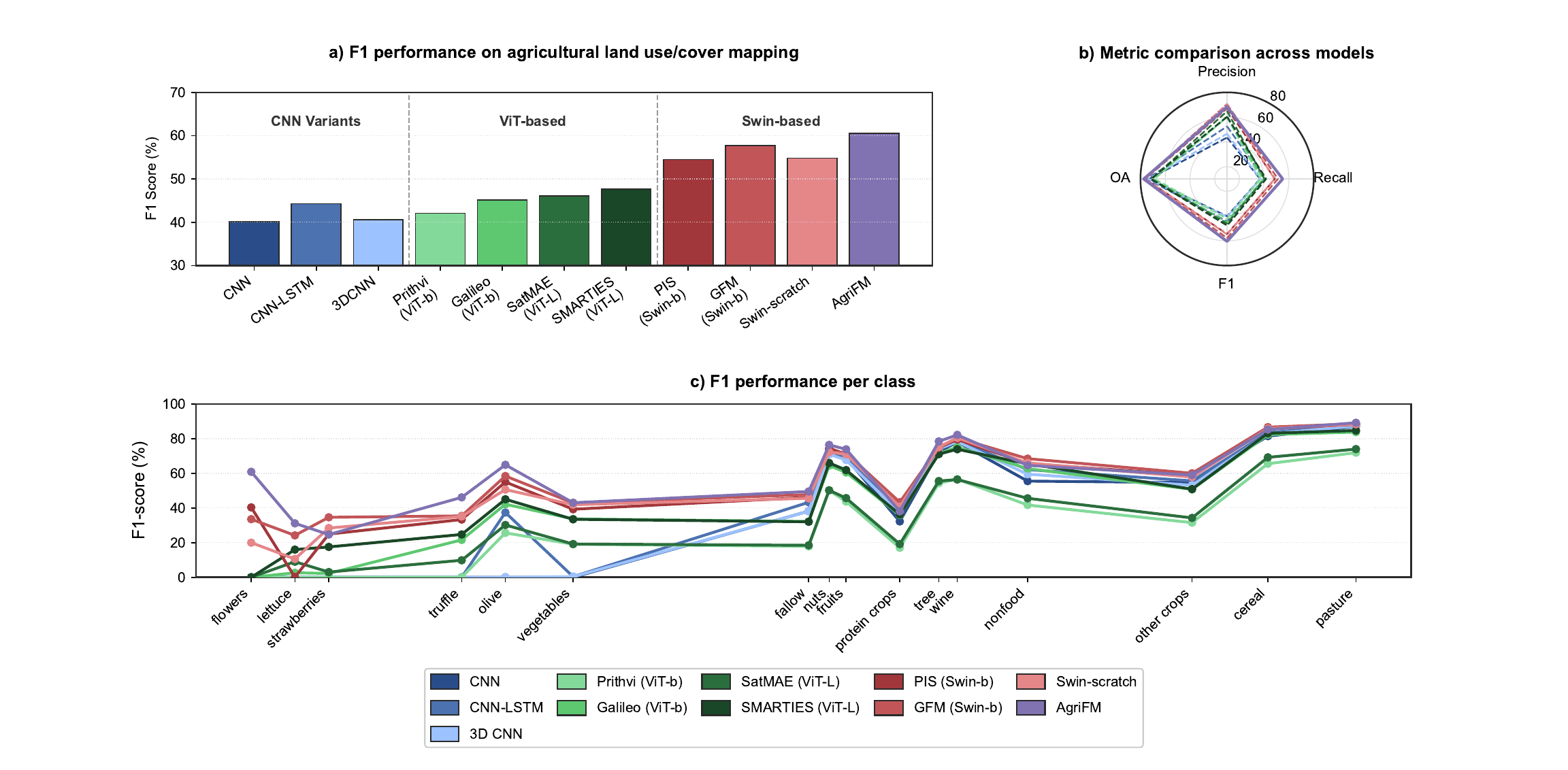}
\caption{{Agricultural land use / land cover mapping  analysis: (a) Method-wise metric comparison, (b) Balanced performance radar chart, (c) Per-class F1 scores ordered by training sample frequency.}}
\label{fig:early_comparison}
\end{figure*}

Spatial results in Figure \ref{fig:early_results} exhibit three characteristic patterns: (1) CNN variants preserve field boundaries but suffer misclassification; (2) ViT methods show superior class discrimination but blurred boundaries and small-field omissions; (3) AgriFM achieves both accurate identification and sharp boundary delineation. These findings collectively validate that hierarchical temporal modeling synergistically optimizes classification accuracy and spatial precision.

Three principal conclusions emerge from the above analysis: First, agricultural land use / land cover demands delicate balance between temporal and spatial feature extraction that conventional single-advantage architectures cannot provide. Second, multi-source temporal pretraining effectively mitigates class imbalance challenges. Third, AgriFM's unified design delivers optimal performance across all scenarios, establishing a new paradigm for precision agriculture. These insights provide both theoretical and practical guidance for agricultural remote sensing model development.

\begin{figure*}
\centering
\includegraphics[width=0.8\linewidth]{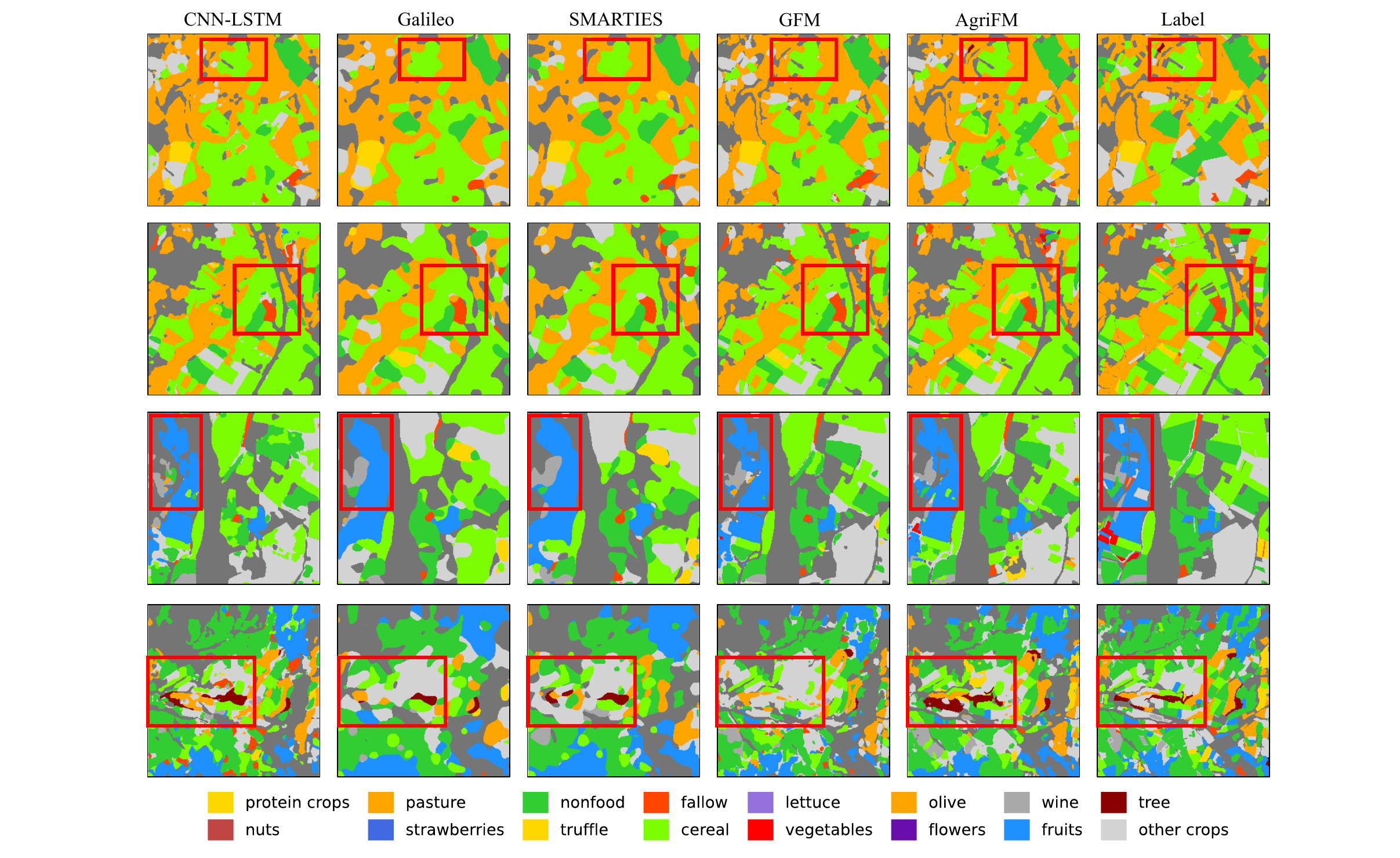}
\caption{{Visual comparison of agricultural land use / land cover mapping  results from test dataset in 2020.  Only  the best-performing methods from each task are shown to ensure clarity in comparison. Different colors in the bottom legend reprensent different land use / land covers. Representative regions with notable differences are highlighted with read bounding boxes for detailed comparison.}}
\label{fig:early_results}
\end{figure*}

We also employ our trained AgriFM to generate a map of agricultural land use / land cover in the ARA region of France for the year 2020, as depicted in Figure \ref{fig:crops_results}. Our model leverages satellite data from the first half of the year, ensuring an accurate mapping of crop distribution in the first half of each year. The map produced by our method effectively reflects the actual crop planting scenario for that year. As can be observed from the figure, there is a high level of crop diversity within the same area, yet our method successfully differentiates between different land use / land covers. 

\begin{figure*}
\centering
\includegraphics[width=0.8\linewidth]{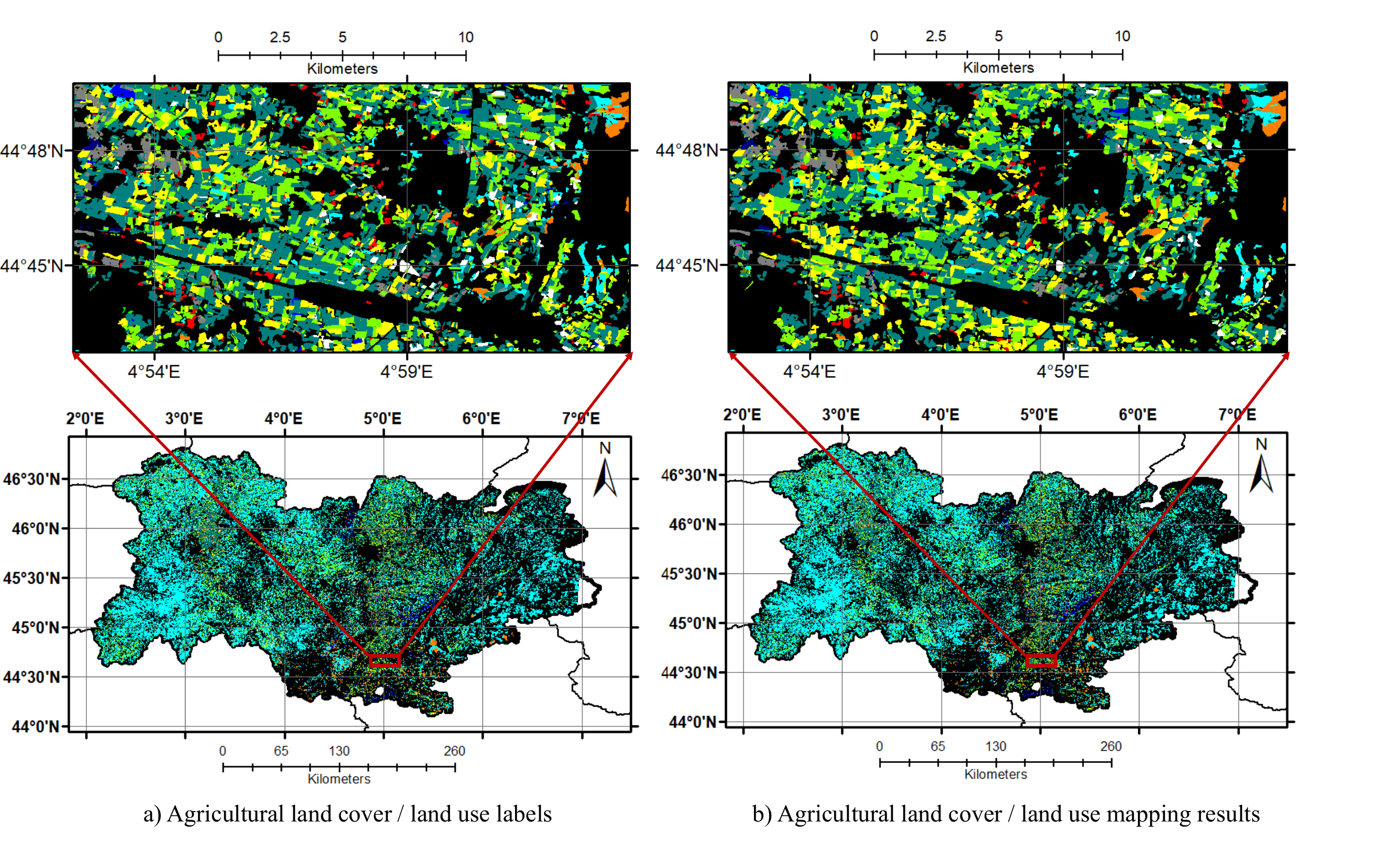}
\caption{Generated map of agricultural land use / land cover mapping in the ARA region of France for the year 2020: a) labels from EuroCrops dataset; b) the mapping results from our method. The lower panel shows the complete mapping coverage across the study area, while the upper panel provides enlarged views of selected representative regions. }
\label{fig:crops_results}
\end{figure*}

\section{Paddy Rice Mapping in Monsoon Asia}

{This experiment evaluates AgriFM's capacity to utilize temporal satellite data for paddy rice mapping in Asia.} We utilize Harmonized Landsat-Sentinel (HLS30) imagery (30m spatial resolution, 6 spectral bands, 5 temporal steps) across Monsoon Asia's rice cultivation belt during 2019. A total of 3,039 training samples, 1,013 validation samples, and 1,013 test samples are drawn from high-confidence rice parcels identified through consensus of existing products. Input sequences span the critical rice-growing period (January–June).

The experimental results in Table \ref{tab:rice_metric} demonstrate the comparative performance of different methods for paddy rice mapping across Monsoon Asia using HLS30 data. All models are tested under consistent conditions. Figure \ref{fig:rice_results} shows some representative rice mapping results.

{Swin-based methods consistently outperform other architectures, with PIS achieving 86.49\% F1 and GFM 86.31\% F1. This advantage over CNN variants (best: CNN at 85.85\% F1) and  ViT models (best: Galileo at 84.79\% F1) confirms the superiority of hierarchical feature extraction for rice mapping. Notably, the randomly initialized Swin-scratch shows competitive performance (85.25\% F1), suggesting the inherent suitability of Swin's architecture for agricultural patterns.}

\begin{table*}[!htb]
    \centering
    \small
    \caption{{Performance comparison on the paddy rice mapping task (\%). The positive column means metrics of paddy rice identification, while the average columns means the average metric of both background and paddy rice.}}
    \begin{tabular}{c|ccc|cccc}
		\toprule
         \multirow{2}{*}{Model} & \multicolumn{3}{c}{Positive (Paddy Rice)} & \multicolumn{4}{c}{Average}\\
        \cmidrule(lr){2-4} \cmidrule(lr){5-8}
         & Precision & Recall & F1 & Precision & Recall & F1 & OA\\
          \midrule
        CNN            & 82.61 & 89.35 & 85.85 & 89.04 & 90.82 & 89.85 & 91.43 \\
        CNN-LSTM       & 81.87 & 88.62 & 85.11 & 88.52 & 90.28 & 89.32 & 90.97 \\
        3DCNN          & 81.48 & 88.38 & 84.79 & 88.27 & 90.07 & 89.08 & 90.77 \\
        Prithvi (ViT-b)& 79.45 & 86.10 & 82.64 & 86.77 & 88.48 & 87.54 & 89.47 \\
        Galileo  (ViT-b) & 82.12 & 87.63 & 84.79 & 88.68 & 90.23 & 89.94 & 91.42\\
        SatMAE (ViT-L) & 80.72 & 85.88 & 83.22 & 87.39 & 88.73 & 88.01 & 89.92 \\
        SMARTIES  (ViT-L)  & 81.37 & 81.01 & 84.10 & 88.18 & 89.76 & 88.92 & 91.02\\
        PIS (Swin-b)   & 83.46 & 89.75 & 86.49 & 89.56 & 91.22 & 90.32 & 91.84 \\
        GFM (Swin-b) & 83.12 & 89.75 & 86.31 & 89.38 & 91.13 & 90.18 & 91.71\\
        AgriFM-scratch  & 80.84 & \textbf{90.17} & 85.25 & 88.30 & 90.70 & 89.35 & 90.92 \\
         AgriFM &  \textbf{84.03} & 90.12 &\textbf{ 86.97} & \textbf{89.93} & \textbf{91.54} & \textbf{90.67} & \textbf{92.14} \\
         \bottomrule
	\end{tabular}
	\label{tab:rice_metric}
\end{table*}

{Further analysis reveals that AgriFM achieves the highest overall performance, with an F1 score of 86.97\% for paddy rice mapping, outperforming all other methods including state-of-the-art Swin-based models like PIS and GFM. This improvement can be attributed to AgriFM's ability to effectively leverage multi-source temporal data and its synchronized spatio-temporal downsampling strategy, which enhances feature representation for complex phenological patterns. Specifically, AgriFM excels in precision (84.03\%) while maintaining high recall (90.12\%), indicating a better balance in minimizing false positives and capturing true rice parcels. The superior performance across all average metrics (precision: 89.93\%, recall: 91.54\%, F1: 90.67\%, OA: 92.14\%) underscores its robustness and generalizability in diverse rice-growing regions. Compared to AgriFM-scratch, which relies solely on architectural advantages, the pre-trained AgriFM demonstrates the added value of land cover fraction supervision and pre-training, leading to more reliable mappings. These results highlight AgriFM's potential as a versatile foundation model for precise agricultural monitoring in challenging, large-scale environments.}

\begin{figure*}
\centering
\includegraphics[width=0.9\linewidth]{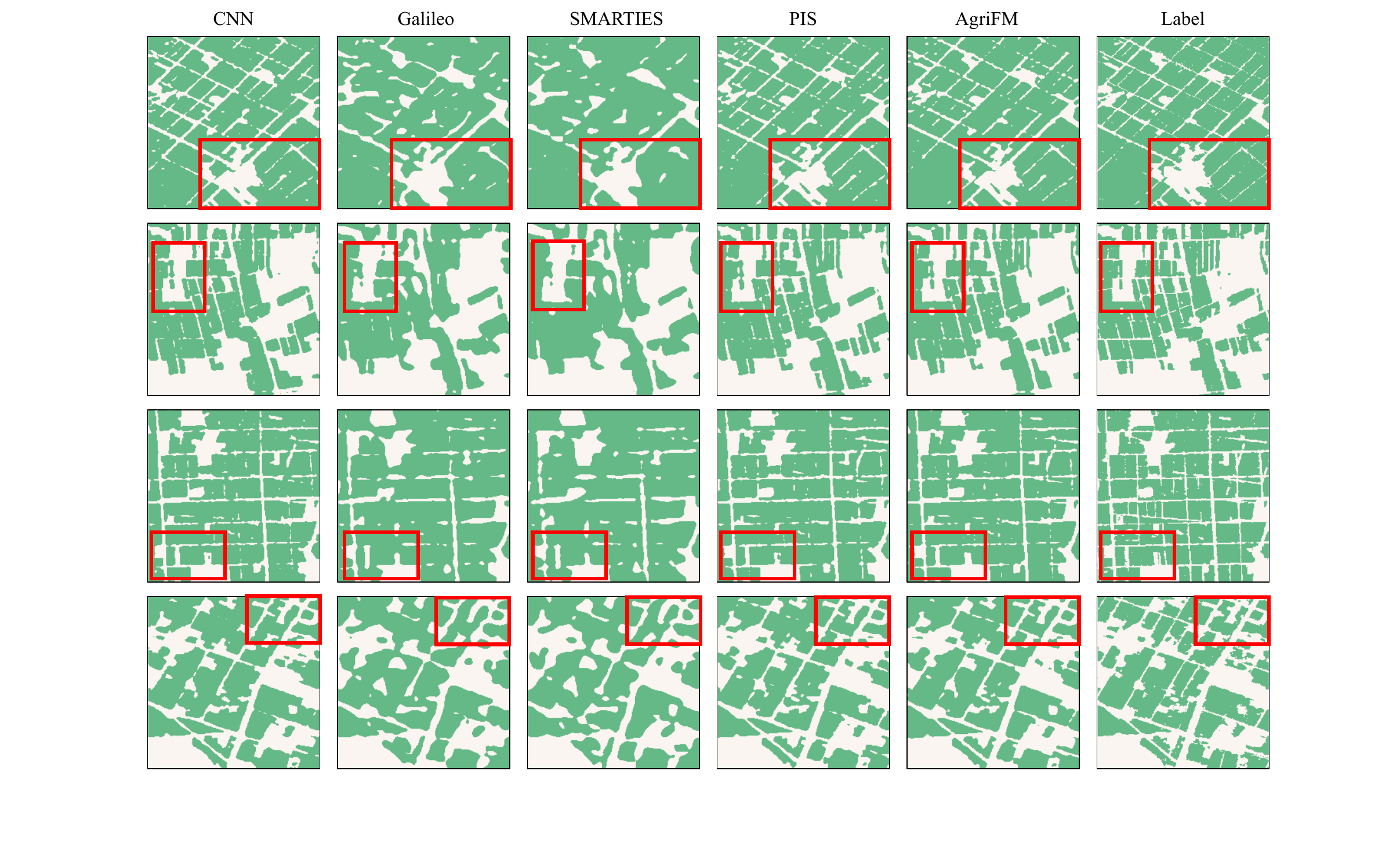}
\caption{{Visual comparison of paddy rice mapping results in Monsoon Asia from test dataset in 2019.  Only  the best-performing methods from each task are shown to ensure clarity in comparison. Paddy rice pixels are shown in green. Representative regions with notable differences are highlighted with red bounding boxes for detailed comparison.} }
\label{fig:rice_results}
\end{figure*}

\section{Winter Wheat Mapping in Asia}
This experiment validates the AgriFM's  performance on Asia's winter wheat mapping. we process MODIS monthly composites (250m\&500m spatial resolution, 7 spectral bands) spanning 11 critical growth stages from pre-sowing (August 2020) to harvest (June 2021). The dataset comprises 2,711 training, 636 validation, and 569 testing tiles (512×512 pixels), geographically stratified to represent diverse agroclimatic zones from the North China Plain to Central Asian steppes.

Table \ref{tab:wheat_metric} presents a comprehensive evaluation of winter wheat mapping across Asia using low-resolution satellite data. Figure \ref{fig:wheat_results} shows some representative visualization for winter wheat mapping.  AgriFM achieves state-of-the-art performance with 75.40\% wheat F1-score and 97.11\% overall accuracy (OA), outperforming Prithvi (66.61\% F1, 95.96\% OA) and SatMAE (67.17\% F1, 96.23\% OA) by significant margins.

{Several key observations emerge from the comparative analysis. First, Swin-based methods consistently dominate the rankings, with PIS (74.47\% F1), GFM (74.33\% F1)  and Swin-scratch (74.23\% F1) already surpassing all CNN and ViT variants. This 4-8\% F1 advantage over ViT models particularly highlights the importance of hierarchical feature extraction when working with low-resolution inputs. Notably, the standard CNN achieves better performance (70.17\% F1) than more complex 3DCNN (67.69\% F1) and CNN-LSTM (68.27\% F1) architectures, suggesting that simple spatial feature extractors may be more effective than sophisticated spatiotemporal designs for this specific task.}

The superior performance of AgriFM can be attributed to two synergistic factors. First, its MODIS-specific pretraining captures unique phenological patterns critical for winter wheat identification - an advantage missing in other models. Second, the architecture's multi-source design enables exceptional feature extraction from low-resolution data, indicating superior detection of true wheat areas. This combination proves especially valuable for winter wheat mapping, where the crop's distinct growth stages  create temporally identifiable signatures in MODIS data.

In addition, while winter wheat's strong phenological characteristics make it relatively distinguishable from other crops, the performance gaps between methods are larger than those observed in rice mapping (Table \ref{tab:rice_metric}). This accentuates our model's specialized capabilities - the 4.4\% F1 improvement over ViT models and 5.68\% over CNNs demonstrate AgriFM's particular aptitude for leveraging subtle temporal cues in coarse-resolution data.

\begin{table*}[!htb]
    \centering
    \small
    \caption{{Performance comparison on the winter wheat mapping task (\%). The positive column means metrics of winter wheat identification, while the average columns means the average metric of both background and winter wheat.}}
    \begin{tabular}{c|ccc|cccc}
		\toprule
         \multirow{2}{*}{Model}  & \multicolumn{3}{c}{Positive (Winter wheat)} & \multicolumn{4}{c}{Average}\\
         \cmidrule(lr){2-4} \cmidrule(lr){5-8} 
         &  Precision & Recall & F1 & Precision & Recall & F1 & OA\\
          \midrule
        CNN            & 73.10 & 67.46 & 70.17 & 85.59 & 82.99 & 84.23 & 96.77 \\
        CNN-LSTM       & 66.53 & 70.11 & 68.27 & 82.37 & 84.00 & 83.16 & 96.34 \\
        3DCNN          & 64.53 & 71.17 & 67.69 & 81.40 & 84.42 & 82.83 & 96.18 \\
        Prithvi (ViT-b)& 62.20 & 71.70 & 66.61 & 80.25 & 84.55 & 82.23 & 95.96 \\
        Galileo  (ViT-b)&  69.12 & 73.95 & 71.45 & 83.56 & 85.71 & 84.59 & 95.81\\
        SatMAE (ViT-L) & 65.76 & 68.64 & 67.17 & 81.94 & 83.26 & 82.58 & 96.23 \\
        SMARTIES  (ViT-L) & 70.41 & 60.95 & 65.34 & 83.73 & 79.50 & 81.44 & 95.41\\
        PIS (Swin-b)   & 73.17 & 75.82 & 74.47 & 85.86 & 87.07 & 86.46 & 97.08 \\
        GFM (Swin-b) & 72.04 & 76.77  & 74.33 & 85.33 & 87.50 & 86.37 & 97.02\\
        Agri-scratch   & 71.78 & \textbf{76.85} & 74.23 & 85.20 & \textbf{87.52} & 86.32 & 97.00 \\
        AgriFM & \textbf{75.56} & 76.14 & \textbf{75.85} & \textbf{87.07} & 87.34 & \textbf{87.20} & \textbf{97.27}\\
         \bottomrule
	\end{tabular}
	\label{tab:wheat_metric}
\end{table*}

\begin{figure*}
\centering
\includegraphics[width=0.9\linewidth]{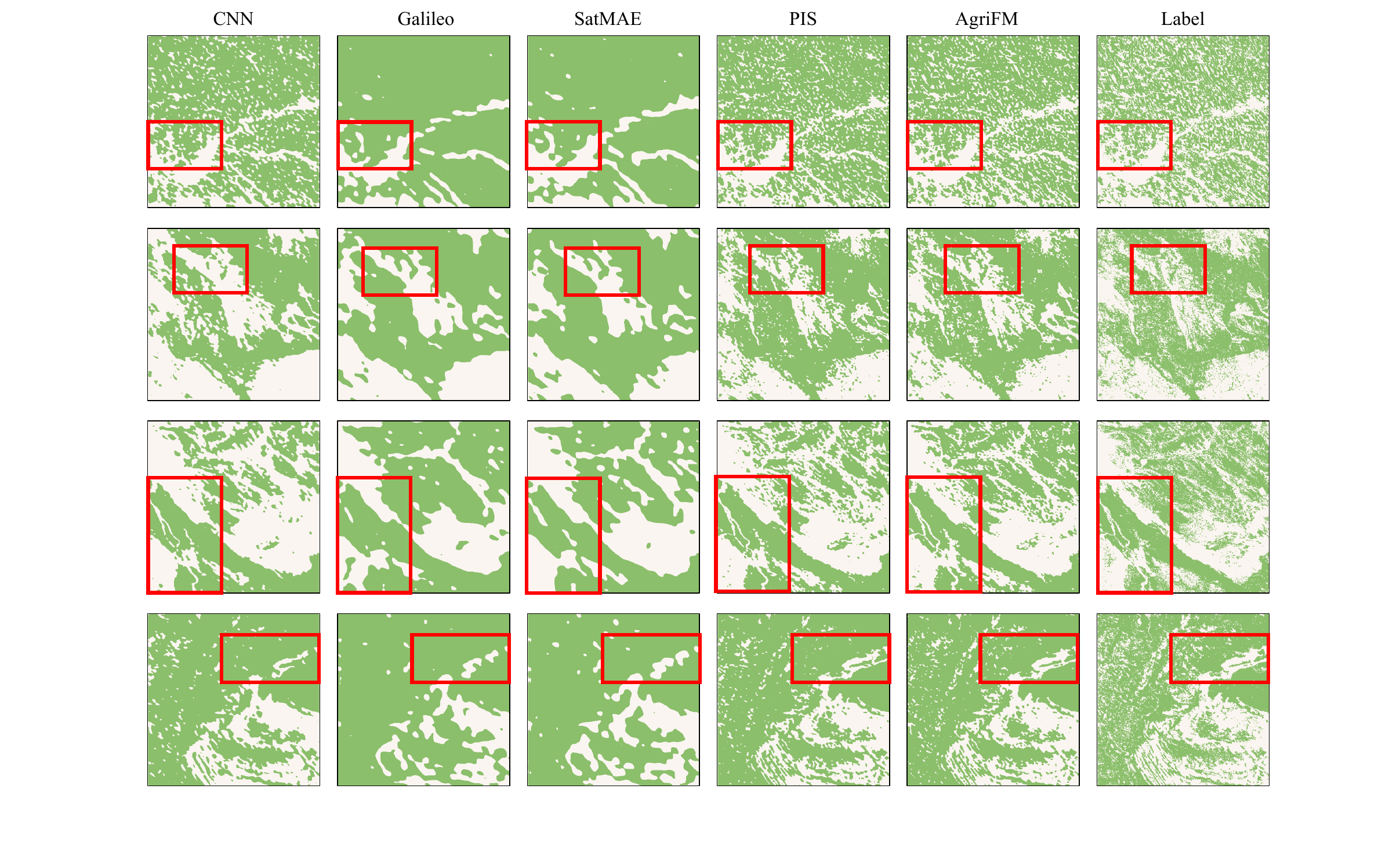}
\caption{{Visual comparison of winter wheat mapping results in Asia from test dataset in 2021.  Only  the best-performing methods from each task are shown to ensure clarity in comparison. Winter wheat pixels are shown in green. Representative regions with notable differences are highlighted with red bounding boxes for detailed comparison.}}
\label{fig:wheat_results}
\end{figure*}

\section{Discussions}

\subsection{{Data-Efficiency Evaluation of Foundation Models with Varying Fine-tuning Samples}}

{To comprehensively evaluate the data efficiency of different foundation models, we conducted extensive experiments across five agricultural mapping tasks with progressively reduced training data ratios (from 100\% to 5\%). This analysis aims to assess the robustness and generalization capability of foundation models when facing limited annotated samples, which is a common challenge in real-world agricultural applications. The experimental setup follows the standard fine-tuning protocol where models pre-trained on large-scale datasets are adapted to downstream tasks with varying amounts of supervised data \citep{pangaea}. Table \ref{tab:scale_exp} presents the detailed performance comparison across all training ratios and tasks, while Figure \ref{fig:scale_exp} illustrates the performance trends through line charts, with the first five subplots corresponding to individual tasks and the final subplot showing the averaged performance across all tasks.}

{The experimental results demonstrate that our proposed AgriFM consistently achieves superior performance across nearly all tasks and training ratios, confirming its exceptional capability in handling data-scarce scenarios. For agricultural land mapping and field boundary delineation tasks, AgriFM maintains a substantial performance advantage throughout all data regimes, with improvements of 2-5\% over the second-best methods. This consistent superiority suggests that the representations learned by AgriFM are particularly well-suited for these fundamental agricultural mapping tasks, even when fine-tuning samples are severely limited.}

{However, a distinct pattern emerges for the more complex agricultural land use/cover mapping task, where all models experience rapid performance degradation as training data decreases. While AgriFM still achieves the best performance in this challenging scenario, the dramatic performance drop across all models underscores that complex fine-grained classification tasks inherently require substantial annotated data to maintain satisfactory performance levels. This observation highlights an important limitation of current foundation models and suggests that for such detailed classification problems, having adequate training samples remains crucial.
}

\begin{table}[!hbt] 
\centering
\small  
\caption{{Performance comparison of foundation models across varying training data ratios (Metrics: mF1 for Agri. land cover / land use mapping, F1 on  positive class for other tasks).  Best performance for each ratio is bolded.}}
\label{tab:scale_exp}
\begin{tabular}{lrrrrrrr} 
\toprule
Model & 100\% & 50.0\% & 33.0\% & 25.0\% & 20.0\% & 10.0\% & 5.0\% \\
\midrule
\multicolumn{8}{c}{\textbf{Agricultural land mapping}} \\  
\midrule

Prithvi     & 73.69 & 69.51 & 69.23 & 69.30 & 66.38 & 66.07 & 63.61 \\
Gelileo     & 75.73 & 75.62 & 73.41 & 68.08 & 69.24 & 68.12 & 65.41 \\
SatMAE      & 76.80 & 75.22 & 74.70 & 72.66 & 70.49 & 69.74 & 67.69 \\
SMARTIES    & 74.76 & 69.26 & 69.60 & 69.63 & 66.91 & 64.63 & 61.16 \\
PIS         & 79.01 & 78.00 & 75.95 & 75.52 & 75.00 & 74.15 & 72.33 \\
GFM         & 81.06 & 77.24 & 75.25 & 75.68 & 75.35 & 74.15 & 72.06 \\
AgriFM      & \textbf{83.09} & \textbf{81.66} & \textbf{80.96} & \textbf{80.57} & \textbf{80.31} & \textbf{76.73} & \textbf{74.27} \\
\midrule
\multicolumn{8}{c}{\textbf{Field boundary delineation}} \\
\midrule
Prithvi     & 53.45 & 55.54 & 49.52 & 56.62 & 56.04 & 54.27 & 49.21 \\
Gelileo     & 62.51 & 56.01 & 56.47 & 55.91 & 55.30 & 55.26 & 52.18 \\
SatMAE      & 62.50 & 59.89 & 57.91 & 55.83 & 54.95 & 53.96 & 52.58 \\
SMARTIES    & 59.92 & 56.86 & 55.98 & 54.95 & 53.90 & 53.02 & 52.89 \\
PIS         & 71.20 & 69.81 & 66.57 & 65.03 & 62.22 & 63.17 & 59.70 \\
GFM         & 72.38 & 68.91 & 66.60 & 65.30 & 64.40 & 62.94 & 62.60 \\
AgriFM      & \textbf{76.27} & \textbf{73.55} & \textbf{71.86} & \textbf{70.83} & \textbf{69.91} & \textbf{68.38} & \textbf{66.36} \\
\midrule
\multicolumn{8}{c}{\textbf{Agricultural land use / land cover mapping}} \\
\midrule
Prithvi     & 42.10 & 37.91 & 35.01 & 35.40 & 34.91 & 30.26 & 25.28 \\
Gelileo     & 45.12 & 40.65 & 40.27 & 37.20 & 35.87 & 32.09 & 26.40 \\
SatMAE      & 46.10 & 42.13 & 41.10 & 37.83 & 35.59 & 31.61 & 25.22 \\
SMARTIES    & 47.65 & 40.48 & 39.13 & 38.30 & 35.64 & 32.02 & 25.42 \\
PIS         & 54.51 & 46.02 & 46.49 & 43.65 & 41.44 & 35.72 & 29.22 \\
GFM         & 57.75 & 49.29 & 47.90 & 44.60 & 42.52 & 35.01 & 30.21 \\
AgriFM      & \textbf{60.49} & \textbf{52.00} & \textbf{49.51} & \textbf{45.67} & \textbf{42.64} & \textbf{35.47} & \textbf{30.33} \\
\midrule

\multicolumn{8}{c}{\textbf{Paddy rice mapping}} \\
\midrule
Prithvi     & 82.64 & 81.61 & 80.74 & 80.10 & 79.99 & 78.92 & 77.01 \\
Gelileo     & 84.79 & 82.27 & 81.44 & 80.90 & 80.75 & 79.17 & 76.87 \\
SatMAE      & 83.22 & 82.28 & 81.30 & 80.46 & 80.70 & 79.00 & 76.68 \\
SMARTIES    & 84.10 & 82.03 & 77.31 & 80.12 & 79.94 & 78.99 & 75.29 \\
PIS         & 86.49 & 85.19 & 84.01 & 83.36 & 83.08 & 81.75 & 78.63 \\
GFM         & 86.31 & 84.76 & 83.61 & 83.36 & 83.09 & 80.92 & 76.90 \\
AgriFM      & \textbf{86.97} & \textbf{86.06} & \textbf{85.31} & \textbf{84.76} & \textbf{84.63} & \textbf{82.70} & \textbf{79.53} \\
\midrule

\multicolumn{8}{c}{\textbf{Winter wheat mapping}} \\
\midrule
Prithvi     & 66.61 & 57.88 & 62.73 & 58.31 & 49.83 & 54.18 & 49.92 \\
Gelileo     & 71.45 & 63.23 & 65.82 & 60.42 & 60.81 & 56.40 & 50.55 \\
SatMAE      & 67.17 & 64.27 & 65.14 & 60.49 & 61.74 & 57.14 & 50.15 \\
SMARTIES    & 65.34 & 57.35 & 58.94 & 57.44 & 56.02 & 45.32 & 36.48 \\
PIS         & 74.47 & 73.34 & 73.93 & 71.55 & 71.49 & 69.17 & 65.80 \\
GFM         & 74.33 & 73.19 & 73.82 & 70.96 & 71.73 & 69.81 & 66.10 \\
AgriFM      & \textbf{75.85} & \textbf{75.17} & \textbf{74.70} & \textbf{73.03} & \textbf{73.81} & \textbf{72.16} & \textbf{68.82} \\
\bottomrule
\end{tabular}
\end{table}

\begin{figure*}
\centering
\includegraphics[width=\linewidth]{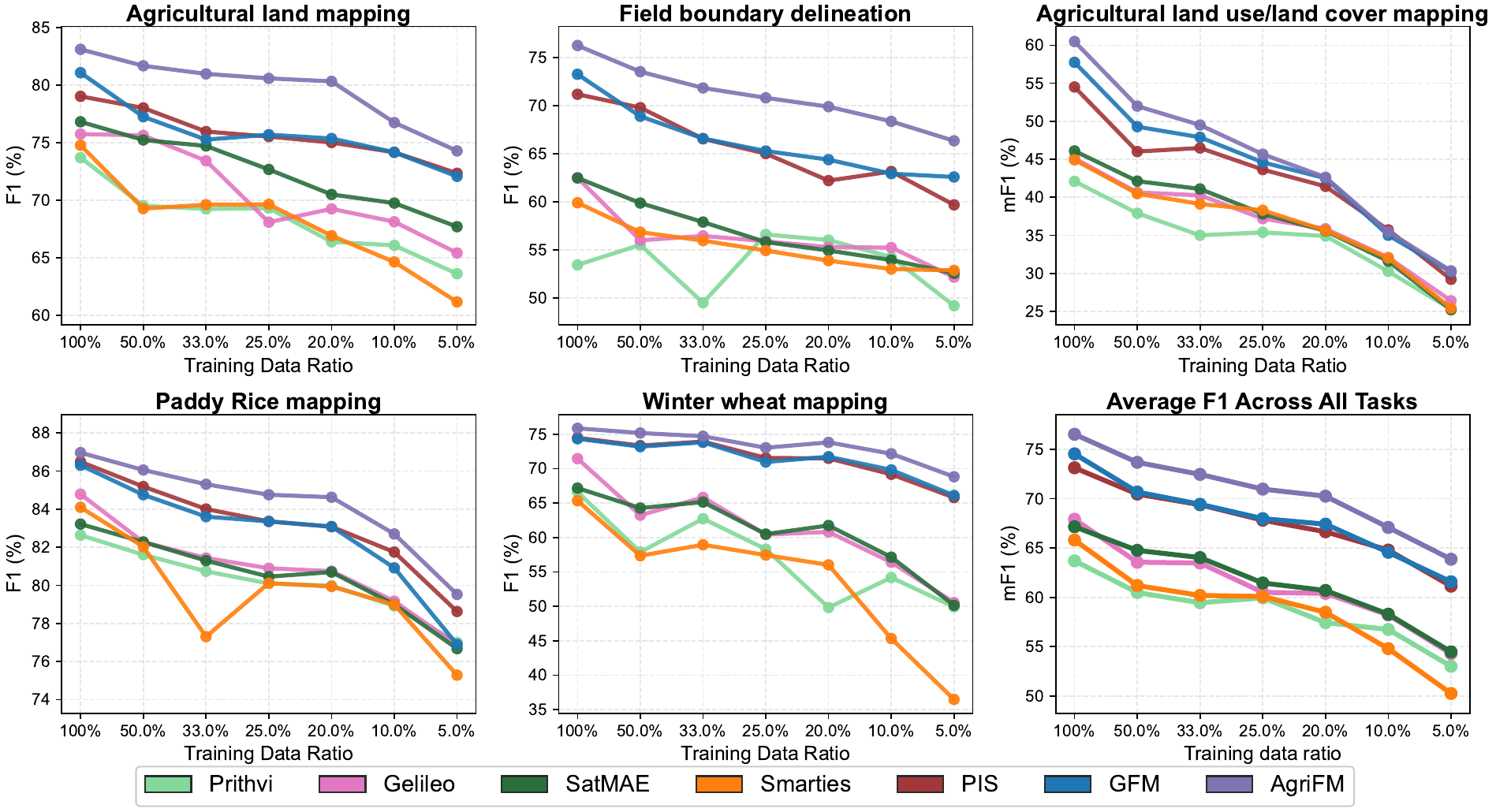}
\caption{{Performance trends of foundation models across different training data ratios. The first five subplots corresponding to individual tasks and the final subplot showing the averaged performance across all tasks.} }
\label{fig:scale_exp}
\end{figure*}

{For specific crop mapping tasks including paddy rice and winter wheat, AgriFM not only achieves the best performance under full data conditions but also demonstrates particularly significant advantages in low-data regimes. The performance gap between AgriFM and other foundation models widens notably when training data is reduced to 20\% or lower, indicating that our model's pre-training strategy effectively captures crop-specific phenological patterns that transfer well even with minimal fine-tuning samples. This characteristic is especially valuable for practical applications where collecting large annotated datasets for specific crops may be challenging or costly.}

{
In summary, these comprehensive experiments validate AgriFM's superior data efficiency and robustness across diverse agricultural mapping tasks. The consistent performance advantage, particularly in low-data scenarios, demonstrates that our method learns more transferable representations during pre-training that require less supervised data for effective adaptation to downstream tasks. These findings position AgriFM as a practical solution for agricultural monitoring applications where annotated data may be limited, while also highlighting the ongoing challenges in complex fine-grained classification scenarios.}

\subsection{{Ablation Study on Key Components of AgriFM}}

{To systematically evaluate the contribution of each component in our proposed AgriFM framework, we conduct comprehensive ablation studies focusing on both pre-training strategies and architectural designs. The experiments are designed to isolate and quantify the impact of four key pre-training elements: (1) the mean-teacher framework for handling label noise in land cover products, (2) land cover fraction supervision during pre-training, (3) multi-source data integration across MODIS, Landsat, and Sentinel-2, and (4) multi-temporal sampling with variable sequence lengths (3-32 frames). Additionally, we analyze the effectiveness of our synchronized temporal downsampling strategy in balancing computational efficiency and performance. For the multi-source ablation, when disabled, we pre-train three separate models using individual data sources and fine-tune each downstream task with the model corresponding to its input data source. Similarly, for the multi-temporal ablation, the disabled setting employs the conventional fixed 16-frame input commonly used in video foundation models.}

{The results presented in Table \ref{tab:pretraining_ablation} demonstrate the progressive improvement achieved by incorporating each component into our framework. Starting from a baseline without mean-teacher and land cover supervision (first row), we observe that adding land cover supervision provides mixed benefits, showing improvements in some tasks while slightly decreasing performance in others. This suggests that while land cover supervision provides valuable semantic guidance, its effectiveness can be limited by label noise in the absence of proper regularization. The incorporation of multi-source data consistently enhances performance across most tasks, particularly benefiting agricultural land mapping and boundary delineation. Notably, the mean-teacher framework emerges as a crucial component, substantially boosting performance across all tasks when combined with other elements. The complete AgriFM configuration with all components achieves the best overall performance, with particularly remarkable improvements in winter wheat mapping (76.65\% F1) and boundary delineation (76.27\% F1), demonstrating the synergistic effect of integrating these complementary design choices.
}

\begin{table}[!htb]
\centering
\caption{{Ablation study on pre-training strategies. MT: Mean-Teacher structure; LC: Land Cover fraction supervision; Multi-source: Using multiple satellite sources during pre-training; Multi-temporal: Variable sequence length input (3-32 frames) during pre-training. Best performance for each task is bolded.}}
\label{tab:pretraining_ablation}
\small
\begin{tabular}{ccccccccc}
\toprule
MT & LC & Multi-source & Multi-temporal & \begin{tabular}{@{}c@{}}Agri. land\\mapping\end{tabular} & \begin{tabular}{@{}c@{}}Boundary\\delineation\end{tabular} & \begin{tabular}{@{}c@{}}Agri. land\\use/cover\end{tabular} & \begin{tabular}{@{}c@{}}Paddy\\rice\end{tabular} & \begin{tabular}{@{}c@{}}Winter\\wheat\end{tabular} \\
\midrule
$\times$ & $\times$ & \checkmark & \checkmark & 81.96 & 73.70 & 59.73 & 86.53 & 73.69 \\
$\times$ & \checkmark & \checkmark & \checkmark & 81.42 & 74.35 & 59.82 & 86.05 & 74.18 \\
$\times$ & \checkmark & $\times$ & \checkmark & 82.70 & 75.22 & 59.93 & 86.09 & 74.08 \\
\checkmark & \checkmark & \checkmark & $\times$ & 81.90 & 73.78 & 59.88 & 86.04 & 75.40 \\
\checkmark & \checkmark & \checkmark & \checkmark & \textbf{83.09} & \textbf{76.27} & \textbf{60.49} & \textbf{86.97} & \textbf{76.65} \\
\bottomrule
\end{tabular}
\end{table}

{Table \ref{tab:downsampling_efficiency} presents a focused analysis of our synchronized temporal downsampling strategy, comparing both performance and computational requirements against a baseline without temporal downsampling. The metrics include: time (total training hours), Mem/b (memory consumption per batch in GB, with batch size indicated after the slash), and FLOPs (floating point operations in GigaFLOPs). While the non-downsampling variant achieves marginally higher performance in agricultural land mapping (83.53\% vs 83.09\%), it suffers significant performance degradation in agricultural land use / land cover mapping (57.58\% vs 60.49\%). More importantly, the computational advantages of our approach are substantial, reducing training time by approximately 40-45\%, memory consumption by 15-33\%, and FLOPs by 60-70\% across different tasks. This efficiency gain is particularly valuable for large-scale agricultural monitoring applications where computational resources are often constrained. }

{These ablation studies collectively demonstrate that each component in AgriFM contributes meaningfully to its overall effectiveness, with the full configuration achieving an optimal balance between performance and efficiency. The synchronized temporal downsampling strategy emerges as particularly crucial, enabling efficient processing of long temporal sequences while maintaining competitive performance across diverse agricultural mapping tasks.}

\begin{table}[!htb]
\centering
\caption{{Performance and computational efficiency of synchronized temporal downsampling. Metrics: time (training hours), Mem/b (memory consumption in GB per batch with batch size), FLOPs (Giga floating-point operations). Best performance for each task is bolded.}}
\label{tab:downsampling_efficiency}
\small
 \resizebox{\linewidth}{!}{
\begin{tabular}{lcccccccccccc}
\toprule
& \multicolumn{4}{c}{\textbf{Agri. land mapping}} & \multicolumn{4}{c}{\textbf{Boundary delineation}} & \multicolumn{4}{c}{\textbf{Agri. land use / land cover}} \\
\cmidrule(lr){2-5} \cmidrule(lr){6-9} \cmidrule(lr){10-13}
Method & F1 & Time & \begin{tabular}{@{}c@{}}Mem/b\\ (GB/bs)\end{tabular} & FLOPs & F1 & Time & \begin{tabular}{@{}c@{}}Mem/b\\ (GB/bs)\end{tabular} & FLOPs & mF1 & Time & \begin{tabular}{@{}c@{}}Mem/b\\ (GB/bs)\end{tabular} & FLOPs \\
\midrule
w/o downsampling & \textbf{83.53} & 17h & 36.2G/2 & 804G & 76.16 & 17h & 36.2G/2 & 580G & 57.58 & 16h & 36.3G/3 & 580G \\
AgriFM & 83.09 & \textbf{10h} & \textbf{30.6G/4} & \textbf{256G} & \textbf{76.27} & \textbf{10h} & \textbf{30.6G/4} & \textbf{256G} & \textbf{60.49} & \textbf{9.5h} & \textbf{24.5G/4} & \textbf{220G} \\
\bottomrule
\end{tabular}
}
\end{table}

\subsection{{ Cross-Temporal and Cross-Source Generalization Analysis}}

{To address the critical aspect of foundation model generalization capabilities, we conducted comprehensive experiments to validate AgriFM's generalization capabilities across temporal dimensions and sources with different spatial resolutions. These experiments assess the model's adaptability to varying input configurations, including different temporal sequence lengths and diverse satellite data sources with varying spatial resolutions. The cross-temporal analysis evaluates performance stability across different temporal contexts, while the cross-spatial analysis examines the model's ability to leverage multi-resolution satellite data for enhanced agriculture mapping.}

{The temporal generalization analysis examines model performance with input sequence lengths ranging from 4 to 24 frames. As demonstrated in Table \ref{tab:temporal_length_analysis} and Figure \ref{fig:temporal_length}, AgriFM consistently outperforms all competing foundation models across all temporal configurations. Notably, while most models exhibit performance degradation with shorter sequences, AgriFM maintains robust performance even with 12-frame inputs (59.00\% mF1), demonstrating only a marginal 1.49\% decrease from the 24-frame configuration. This temporal robustness stems from our synchronized spatio-temporal downsampling strategy, which enables effective feature extraction regardless of input sequence length.}

\begin{table}[!htb]
\centering
\caption{{Performance comparison (mF1 scores) on agricultural land use/cover mapping with varying input sequence lengths (4, 8, 12, 16, 24 frames). Best performance for each sequence length is bolded.}}
\label{tab:temporal_length_analysis}
\small
\begin{tabular}{lccccc}
\toprule
Model & 24 frames & 16 frames & 12 frames & 8 frames & 4 frames \\
\midrule
Prithvi & 42.10 & 43.01 & 41.60 & 38.93 & 39.85 \\
Gelileo & 45.12 & 43.52 & 42.60 & 41.81 & 39.75 \\
SatMAE & 46.10 & 45.34 & 44.70 & 43.49 & 39.69 \\
SMARTIES & 47.65 & 44.15 & 43.20 & 41.52 & 39.50 \\
PIS & 54.51 & 53.32 & 49.40 & 47.13 & 44.11 \\
GFM & 57.75 & 53.23 & 53.00 & 48.46 & 46.24 \\
AgriFM & \textbf{60.49} & \textbf{59.87} & \textbf{59.00} & \textbf{53.62} & \textbf{48.08} \\
\bottomrule
\end{tabular}
\end{table}

\begin{figure*}
\centering
\includegraphics[width=\linewidth]{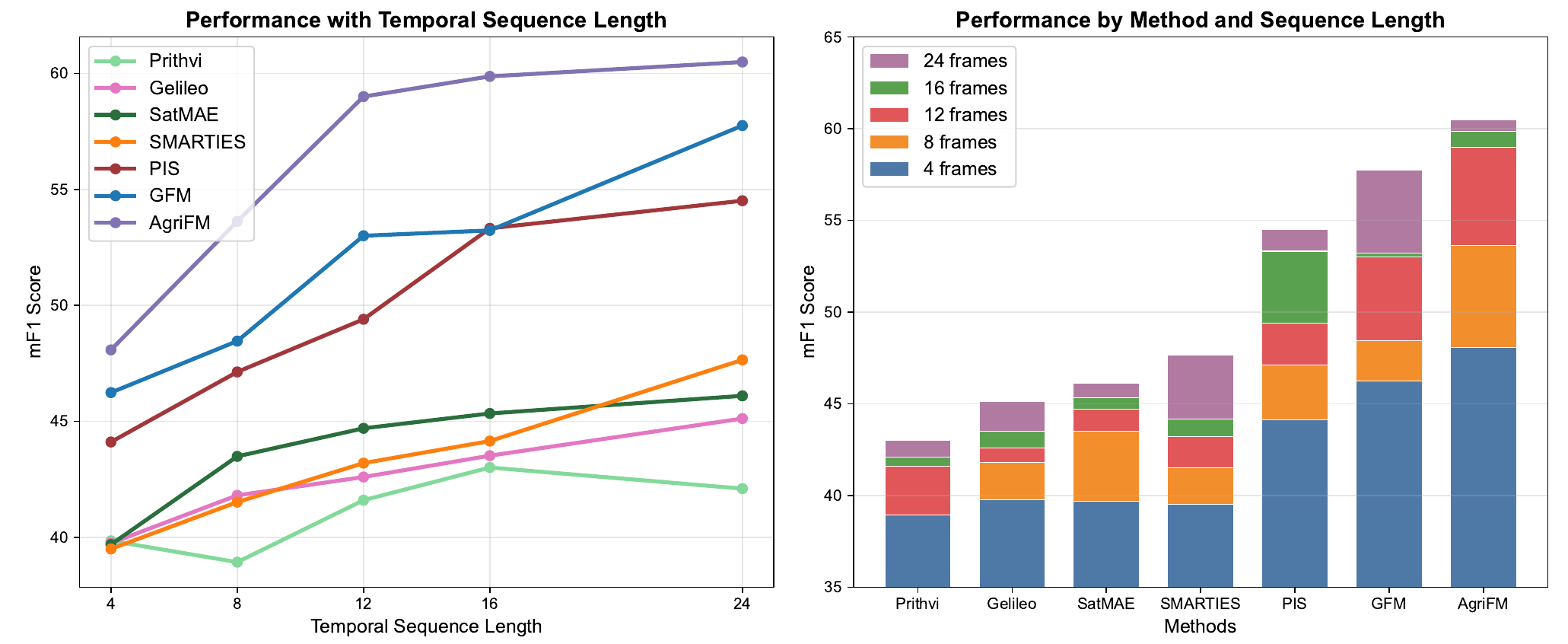}
\caption{{Performance comparison (mF1 scores) on agricultural land use/cover mapping with varying input sequence lengths (4, 8, 12, 16, 24 frames).}} 
\label{fig:temporal_length}
\end{figure*}

{
The cross-source generalization analysis, presented in Table \ref{tab:cross_spatial_generalization}, reveals insightful patterns about AgriFM's ability to leverage diverse satellite data sources. Sentinel-2 data (10m resolution) serves as our high-resolution baseline, providing the highest individual performance across most tasks due to its superior spatial and temporal resolution. However, its temporal coverage is limited to post-2017 data. Landsat 8/9 data (30m resolution) with 12 monthly frames shows consistently lower performance than Sentinel-2, with particularly pronounced degradation in complex tasks like agricultural land use/cover mapping (33.59\% vs 60.49\% mF1).}

{
The integration of MODIS data with its comprehensive 44-frame annual coverage demonstrates complementary benefits. For cropland mapping, the Landsat+MODIS combination (83.14\% F1) surpasses Sentinel-2 alone (83.09\% F1), with similar improvements observed across other tasks. This suggests that MODIS's dense temporal sampling provides valuable phenological information that compensates for its coarser spatial resolution. Furthermore, the fusion of Sentinel-2 and Landsat data yields substantial improvements in agricultural land mapping (85.61\% vs 83.09\% F1) and field boundary delineation (78.39\% vs 76.27\% F1), indicating that the complementary characteristics of these sensors enhance spatial feature extraction.
}

{
However, for the more complex agricultural land use/cover mapping task, the benefits of data fusion are more nuanced. While Sentinel-2 alone achieves 60.49\% mF1, the addition of Landsat data provides only marginal improvement (60.57\% mF1), suggesting that Landsat's limited additional information has minimal impact on fine-grained classification performance. This pattern highlights that while multi-source integration generally benefits agricultural mapping tasks, its effectiveness is task-dependent, with complex classification problems requiring the high-quality features provided by premium data sources like Sentinel-2.
}

{
These comprehensive experiments demonstrate that AgriFM possesses strong cross-temporal and cross-source generalization capabilities, effectively leveraging diverse temporal sequences and multi-resolution satellite data to enhance agricultural mapping performance across various tasks and configurations.}

\begin{table}[h!]
\centering
\caption{{Performance across different satellite data sources with varying spatial resolutions (Sentinel-2: 10m\&20m; Landsat 8/9: 30m; MODIS: 250m\&500m) and temporal configurations. Numbers in data source columns indicate temporal frame counts.}}
\label{tab:cross_spatial_generalization}
\small
\begin{tabular}{c|ccc|ccc}
\toprule
Task & \begin{tabular}{@{}c@{}}Sentinel-2\\frames\end{tabular} & \begin{tabular}{@{}c@{}}Landsat\\frames\end{tabular} & \begin{tabular}{@{}c@{}}MODIS\\frames\end{tabular} & Precision & Recall & F1 \\
\midrule

\multirow{4}{*}{Agri. land mapping}& 32 & -- & -- & 84.31 & 81.90 & 83.09 \\
& -- & 12 & -- & 83.54 & 81.78 & 82.65 \\
& 32 & 12 & -- & \textbf{86.18} & \textbf{85.05} & \textbf{85.61} \\
& -- & 12 & 44 & 71.74 & 84.58 & 83.14 \\
\midrule
\multirow{4}{*}{Field boundary delineation} 
& 32 & -- & -- & 75.11 & 77.47 & 76.27 \\
& -- & 12 & -- & 72.78 & 73.45 & 73.11 \\
& 32 & 12 & -- & \textbf{77.91} & \textbf{78.88} & \textbf{78.39} \\
& -- & 12 & 44 & 73.51 & 74.14 & 73.82 \\
\midrule
\multirow{4}{*}{Agri. land use/cover mapping} 
& 24 & -- & -- & 68.34 & 54.97 & 60.49 \\
& -- & 12 & -- & 49.22 & 27.86 & 33.59 \\
& 24 & 12 & -- & \textbf{68.37} & \textbf{55.08} & \textbf{60.57} \\
& -- & 12 & 44 & 55.26 & 28.55 & 34.95 \\
\midrule
\multirow{2}{*}{Paddy rice mapping} 
& -- & 5 & -- & \textbf{84.03} & 90.12 & 86.97 \\
& -- & 5 & 44 & 83.61 & \textbf{91.63} & \textbf{87.44} \\
\bottomrule
\end{tabular}
\end{table}

\subsection{{Limitation and future directions}}

{While the proposed AgriFM demonstrates strong performance across multiple agricultural mapping tasks, we acknowledge several limitations that point toward valuable future research directions. This part focuses on two primary aspects: the data alignment strategy employed during pre-training and the generalizability beyond the current application scope.}

{\textbf{Data alignment strategy:} Our method employs a non-paired multi-source learning strategy, which differs from conventional approaches that require  spatial alignment across different satellite sensors. This design choice is supported by the use of land cover fraction supervision, which serves as a semantic bridge that enables feature-space alignment across varying spatial resolutions and sensor characteristics. The land cover fractions provide a consistent learning target that transcends the specific geometric properties of individual sensors, allowing the model to learn robust representations that are invariant to the particularities of each data source.}
    
{ While our non-paired approach has demonstrated effectiveness in learning transferable representations, future work could explore the complementary benefits of paired multi-source data. A carefully constructed fully-aligned dataset with land cover supervision would enable more direct comparisons between alignment strategies and potentially reveal additional insights about cross-sensor relationships. Such investigations could further advance our understanding of optimal strategies for integrating heterogeneous remote sensing data in foundation model development.}

{\textbf{Versatility of decoder:} The current architecture of decoder, though unified across different mapping tasks, still requires task-specific fine-tuning and may not generalize effectively to fundamentally different problem paradigms beyond dense spatial prediction. Specifically, our decoder is not optimized for tasks such as site-based regression, point prediction, or other non-spatial learning objectives that diverge from the pixel-wise mapping framework. This limitation stems from our design focus on addressing the specific challenges of agricultural land monitoring, where the output consistently takes the form of spatial maps derived from satellite image sequences.}
    
{Looking forward, several promising directions emerge for enhancing decoder generalizability. Future work could explore the development of truly task-agnostic decoders through multi-task learning frameworks that simultaneously handle diverse agricultural monitoring objectives. Additionally, investigating dynamic decoder architectures that can adapt their structure based on task requirements presents an exciting research avenue. The integration of prompt-based mechanisms or task-conditioning approaches could further improve the decoder's flexibility across different agricultural applications. Another valuable direction involves extending the decoder's capabilities beyond traditional classification tasks to encompass regression-based agricultural monitoring, such as yield prediction or biophysical parameter estimation, while maintaining the architectural efficiency that characterizes our current design.}
    
{\textbf{Spatial generalization of remote sensing foundation models:} The geographical scope of our pre-training data remains limited compared to the global coverage of modern satellite systems. The fundamental question of how to optimize pre-training strategies for maximum geographical transferability needs deeper investigation. Unlike other domains where data can be considered independent samples, remote sensing observations are intrinsically spatial-dependent and constrained to Earth's surface. This unique characteristic raises important questions about whether including specific geographical regions during pre-training systematically improves downstream performance in those same region.}

{The development of truly global foundation models presents both unprecedented opportunities and substantial challenges. The construction of a comprehensive global temporal dataset would require  huge storage and  computing capabilities. However, such investment could yield significant returns, potentially enabling robust zero-shot generalization through integration with multi-modal data sources like existing global products and language models. The parallel with scaling laws in language models suggests that increased data coverage and model capacity may systematically improve performance, though the unique spatial constraints of geographical data may introduce fundamentally different scaling behaviors that merit dedicated investigation.}

\section{Conclusion}

This paper first addresses and validates the necessity of simultaneous hierarchical
spatiotemporal feature extraction for crop  mapping. It leads to the development of a modified Video Swin Transformer  architecture where temporal down-sampling is synchronized with spatial down-sampling operations. Capitalizing on this finding, we introduce AgriFM, a multi-source temporal remote sensing foundation model with specialized spatiotemporal modeling for agricultural crop  mapping. AgriFM integrates multi-source satellite data including MODIS, Landsat-8/9, and Sentinel-2, and employs geographical land cover products for supervised pre-training. It effectively addresses the limitations of existing RSFMs, ensuring the incorporation of multi-resolution data, providing comprehensive coverage of the crop growth cycle, optimizing the use of geographical information, and offering a unified framework for a wide array of crop mapping tasks. These tasks include, but are not limited to, agricultral land mapping, field boundary delineation, and specific tasks like winter wheat and rice mapping.
The effectiveness and adaptability of AgriFM have been substantiated through superior performance across a variety of crop mapping tasks, demonstrating its potential as an immediately applicable solution for leveraging deep learning in more precise and efficient crop mapping.

The significance of this method lies not only in its immediate practical application but also in its potential for future advancements in this critical field. With the escalating global population growth and climate change, food security has become a pressing issue requiring efficient crop mapping, and satellite remote sensing plays a pivotal role in this regard. However, the challenges presented by current methods necessitate the development of more efficient and precise models such as AgriFM. By addressing the limitations of existing RSFMs and providing a comprehensive framework for diverse tasks, AgriFM paves the way for future advancements in crop mapping, contributing significantly to the efforts towards achieving global food security.

Despite the agricultural applications, the natural extension of this work lies in assessing the model's generalizability to other Earth observation tasks. Many geo-spatial applications similarly require simultaneous processing of spatiotemporal data, suggesting our foundation model could serve as a versatile backbone. However, task-specific adaptations—particularly in decoder design—remain necessary. For vegetation parameter estimation (e.g., LAI and FVC), for instance, the decoder must generate time-specific outputs rather than integrated predictions. While Swin Transformer outperforms other architectures for crop mapping, performance variations across tasks are substantial. For certain image classification or regression tasks, ViT-based methods may still prevail due to their global representation capabilities.

Our results confirm that pretraining significantly boosts performance, especially for complex tasks. However, architectural suitability remains equally crucial. This study's practical implication is clear: for agriculture mapping tasks, Video Swin Transformer should be the architecture of first consideration, with domain-specific pretraining then applied for additional gains. This two-stage approach—selecting the optimal architecture followed by targeted pretraining—proves more effective than relying solely on either component.

Regarding pretraining strategies, while masked image modeling and contrastive learning dominate current practice, our choice of land cover supervision offers distinct advantages. Existing studies have demonstrated the value of incorporating remote sensing-specific knowledge into pretraining. As a fundamental geographic prior, land cover information provides rich semantic guidance. This raises intriguing possibilities—many other Earth observation products (e.g., biomass, soil moisture) could theoretically serve as pretraining supervision, though their effectiveness requires systematic evaluation.

\section*{Data Availability}
Codes and models are available on the website of the Jockey Club STEM Lab of Quantitative Remote Sensing, HKU (\url{https://glass.hku.hk}) and \url{https://github.com/flyakon/AgriFM}.

\section*{Acknowledgment}
The research work described in this paper was conducted in the JC STEM Lab of Quantitative Remote Sensing funded by The Hong Kong Jockey Club Charities Trust. 

The computations were performed using research computing facilities offered by Information Technology Services, the University of Hong Kong.

\bibliographystyle{elsarticle-num-names}

\bibliography{refbib}
\end{document}